\documentclass[journal, twoside]{IEEEtran} % for RA-L (camera ready only!), T-RO, etc.
\usepackage{cite}
\usepackage{amsmath,amssymb,amsfonts}
\usepackage[pdftex]{graphicx} % pdftex option added by Freek
\usepackage{textcomp} % The package supports the Text Companion fonts
\usepackage{xcolor}
\def\BibTeX{{\rm B\kern-.05em{\sc i\kern-.025em b}\kern-.08em
    T\kern-.1667em\lower.7ex\hbox{E}\kern-.125emX}}

\usepackage{hyperref}

\usepackage{cleveref}
\Crefname{equation}{Eq.}{Eqs.} % Ensure abbreviation used for capital version also.
\Crefname{figure}{Fig.}{Figs.}

\graphicspath{{images-src/}{images-bin/}} % Read from both source and compiled image directories. 

\definecolor{dlrprim1}{HTML}{000000} 
\definecolor{dlrprim2}{HTML}{666666} 
\definecolor{dlrprim3}{HTML}{b9cad2}
\definecolor{dlrprim4}{HTML}{ffffff} 

\definecolor{dlrblue1}{HTML}{00658b} 
\definecolor{dlrblue2}{HTML}{3b98cb}
\definecolor{dlrblue3}{HTML}{6cb9dc}
\definecolor{dlrblue4}{HTML}{a7d3ec}
\definecolor{dlrblue5}{HTML}{d1e8fa}

\definecolor{dlryellow1}{HTML}{d2ae3d}  
\definecolor{dlryellow2}{HTML}{f2cd51} 
\definecolor{dlryellow3}{HTML}{f8de53}
\definecolor{dlryellow3}{HTML}{fcea7a}
\definecolor{dlryellow3}{HTML}{fff8be}

\definecolor{dlrgreen1}{HTML}{82a043} 
\definecolor{dlrgreen2}{HTML}{a6bf51}
\definecolor{dlrgreen3}{HTML}{cad55c}
\definecolor{dlrgreen4}{HTML}{d9df78}
\definecolor{dlrgreen5}{HTML}{e6eaaf}

\definecolor{dlrgray1}{HTML}{666666} 
\definecolor{dlrgray2}{HTML}{868585}
\definecolor{dlrgray3}{HTML}{b1b1b1}
\definecolor{dlrgray4}{HTML}{cfcfcf}
\definecolor{dlrgray5}{HTML}{ebebeb}

\usepackage[textsize=scriptsize]{todonotes} 
\usetikzlibrary{arrows}
\usetikzlibrary{patterns.meta}
\usetikzlibrary{decorations.pathmorphing}
\usetikzlibrary{calc}

\usepackage{subcaption}
\usetikzlibrary{tikzmark, decorations.pathreplacing, calligraphy}

\usepackage{booktabs}
\usepackage{colortbl}
\definecolor{Light0}{rgb}{0.95, 1, 0.95}
\definecolor{Light1}{rgb}{0.98, 0.95, 0.90}
\definecolor{Light2}{rgb}{0.98, 0.98, 0.93}
\definecolor{Light3}{rgb}{0.98, 0.98, 1}
\definecolor{Light4}{rgb}{0.93, 0.98, 0.98}

\usepackage[acronym]{glossaries}
\glsdisablehyper
\newacronym{vf}{VF}{Virtual Fixture}
\newacronym{gmm}{GMM}{Gaussian Mixture Model}
\newacronym{gmr}{GMR}{Gaussian Mixture Regression}
\newacronym[longplural=Gaussian Processes]{gp}{GP}{Gaussian Process}
\newacronym[longplural=degrees of freedom]{dof}{DoF}{degree of freedom}

\usepackage{bm}
\usepackage[separate-uncertainty=true]{siunitx}
\usepackage{algorithm}
\usepackage{algpseudocode}

\newcommand{\rmcaffiliation}{German Aerospace Center (DLR), Robotics and Mechatronics Center (RMC), M\"unchener Str. 20, 82234 We\ss ling, Germany.}
\newcommand{\tumaffiliation}{School of Computation, Information and Technology, Sensor Based Robotic Systems and Intelligent Assistance Systems, Technical University Munich, Friedrich-Ludwig-Bauer-Str. 3, Garching, Germany.}

\IEEEoverridecommandlockouts                             

\title{Passive Variable Impedance For Shared Control}

\author{Maximilian Mühlbauer$^{1,2}$, Nepomuk Werner$^{2, 1}$, Ribin Balachandran$^2$, Thomas Hulin$^2$, João Silvério$^2$, Freek Stulp$^2$, Alin Albu-Schäffer$^{2, 1}$
\thanks{$^1$ \tumaffiliation}  % Affiliation defined above.
\thanks{$^2$ \rmcaffiliation}  % Affiliation defined above.
}

\begin{document}

\maketitle

%beginispell

%%%%%%%%%%%%%%%%%%%%%%%%%%%%%%%%%%%%%%%%%%%%%%%%%%%%%%%%%%%%%%%%%%%%%%%%%%%%%%%%
\begin{abstract}
Shared Control methods often use impedance control to track target poses in a robotic manipulator.
The guidance behavior of such controllers is shaped by the used stiffness gains, which can be varying over time to achieve an adaptive guiding.
When multiple target poses are tracked at the same time with varying importance, the corresponding output wrenches have to be arbitrated with weightings changing over time.
In this work, we study the stabilization of both variable stiffness in impedance control as well as the arbitration of different controllers through a scaled addition of their output wrenches, reformulating both into a holistic framework.
We identify passivity violations in the closed loop system and provide methods to passivate the system.
The resulting approach can be used to stabilize standard impedance controllers, allowing for the development of novel and flexible shared control methods.
We do not constrain the design of stiffness matrices or arbitration factors; both can be matrix-valued including off-diagonal elements and change arbitrarily over time.
The proposed methods are furthermore validated in simulation and in real robot experiments on different systems, proving their effectiveness, comparing favorably to baselines and showcasing different behaviors which can be utilized depending on the requirements of the shared control approach.
\end{abstract}

%\begin{IEEEkeywords}
%component, formatting, style, styling, insert
%\end{IEEEkeywords}

%%%%%%%%%%%%%%%%%%%%%%%%%%%%%%%%%%%%%%%%%%%%%%%%%%%%%%%%%%%%%%%%%%%%%%%%%%%%%%%%
\section{Introduction}
\label{sec:introduction}
% variable impedance overview
Impedance control \cite{hogan1984impedance} is often employed in shared control methods as it allows for a compliant interaction with the environment as well as the human operator.
A high stiffness along certain \glspl{dof} leads to a strong guidance, while a low stiffness allows for more compliance for the operator command.
Those \glspl{dof} do not necessarily need to align with coordinate axes, thus resulting in off-diagonal stiffness matrix entries~\cite{abifarraj2017learning} or even couplings between position and orientation~\cite{muehlbauer2025unified}.
Also, different stiffness levels can be interpreted as different levels of haptic authority \cite{abbink2011haptic}.
Learning and control of variable impedance behaviors have therefore recently gained traction with an overview given by \cite{abudakka2020variable}.

\begin{figure}
	\centering
    \includegraphics[width=\columnwidth,trim={400 170 50 10},clip,page=1]{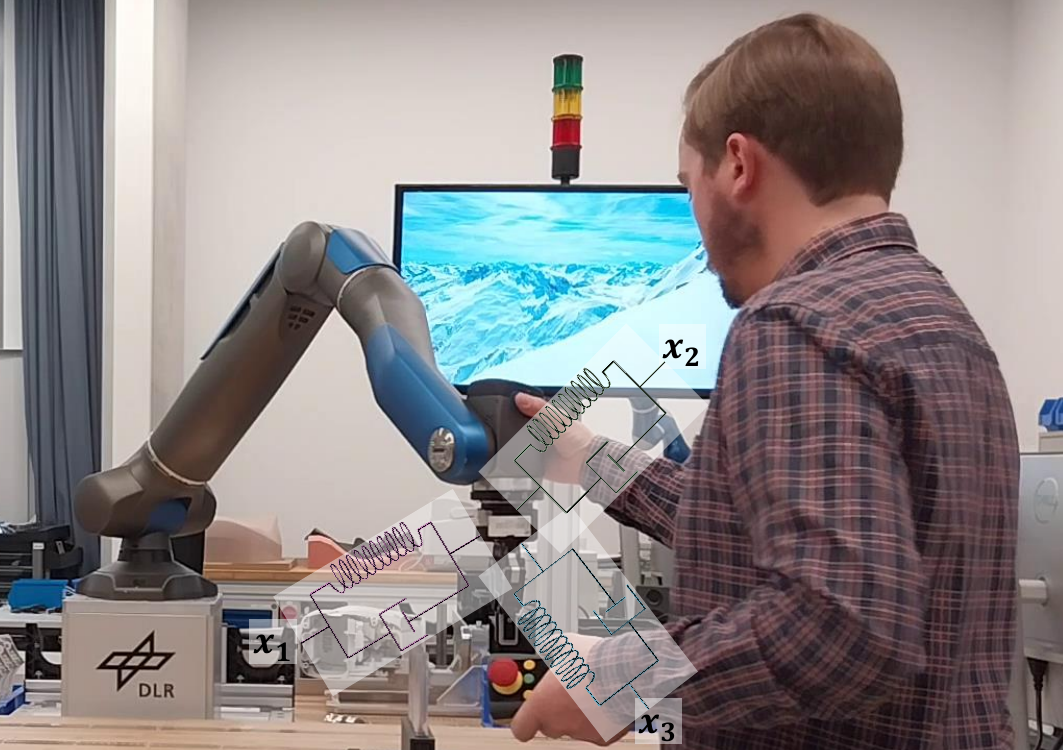} % page 1 is non-anon, page 2 is anon
	\caption{\label{fig:intro:overview} Multiple impedance-based shared control methods with attractors $\bm{x}_1$, $\bm{x}_2$ and $\bm{x}_3$ acting in parallel on a robotic manipulator. We provide methods for the passivation of variable stiffness and arbitration gains for these controllers.\vspace{-1em}}
\end{figure}

% some variable impedance learning approaches, stability
Shaping the impedance behavior is, for example, possible by leveraging the uncertainty of probabilistic machine learning methods.
For instance, a \gls{gp} can be used to learn variable impedance behavior \cite{franzese2021ilosa}.
Other approaches \cite{abifarraj2017learning,muehlbauer2024probabilistic,muehlbauer2025unified} utilize the uncertainty of attractors to modulate the stiffness.
This method results in fully populated, time-varying stiffness matrices, which can therefore also model couplings between positional and rotational \glspl{dof}.
%\todo{also cite \cite{silverio2019uncertainty}}
Furthermore, multiple attractor points regulated through such time-varying stiffness are arbitrated using time-varying gains.
These approaches however do not prove stability of the resulting controller.

In \cite{michel2023learning}, a \gls{gmm} coupled with \gls{gmr} is used for learning the stiffness of attractor poses, with an energy tank to ensure passivity.
Previously dissipated energy is also used by~\cite{ferraguti2013tank} to stabilize passivity-violating stiffness increases.
% stability without energy tank
Contrarily, we propose a novel approach that does not rely on storing dissipated energy but utilizes the power available in the current time step to perform stiffness changes that would otherwise violate passivity.
This design choice ensures that the robot remains stable and predictable, as power input only comes from the user for stationary target poses.
Time-domain passivity~\cite{hannaford2002time} would similarly ensure the dissipation of passivity-violating power at each time step.
To achieve such dissipation, however, either position drift or force jitter result.

The stability of a variable stiffness time series can also be analyzed \cite{kronander2016stability} if the trajectory is known beforehand, which is not the case in shared control.
With passivity filters \cite{bednarczyk2020passivity}, this approach is generalized to stiffness values not known beforehand.
The resulting approach however only supports diagonal stiffness matrices.
Furthermore, diagonal mass matrices are assumed, which is violated for task-space control of robots.

Beyond variable impedance, the problem of \textit{arbitration} between different controllers often arises in shared control.
Conceptually closest, but only valid for scalar factors, are \cite{balachandran2021finite} and \cite{balachandran2023passive}.
While the first one achieves a passivation of the arbitration of wrenches with scalar factors, however, at the cost of a position drift, the latter stabilizes the scaling of stiffness matrices using a scalar factor.

% our solution
In our work, we show how passivity-based formulations can be used to stabilize time-varying matrix-valued scalings $\bm{S}_i$ as well as time-varying fully populated stiffness matrices $\bm{K}_{\mathrm{p},i}$.
This allows us to passivate shared control approaches such as \cite{muehlbauer2025unified} without imposing requirements on the scaling factors or stiffness matrix changes and without position drift, thus giving maximum design flexibility to the developers of shared control algorithms.
The contributions of this work are:
\begin{enumerate}
	\item a novel passivation of variable impedance, limiting spring deflection with lever scaling (\Cref{sec:stabilization:spring_deflection}),
	\item an extension of the passivation by limiting the stiffness change from scalar factors \cite{balachandran2023passive} to arbitrarily varying fully populated stiffness matrices (\Cref{sec:stabilization:stiffness_change}),
	\item a reformulation of shared control arbitration from a weighted addition \eqref{eq:impedance_to_torque} to a modulation of the stiffness used in the impedance controller (\Cref{sec:problem}),
	\item an evaluation and comparison to baselines (\Cref{sec:experiments}).
\end{enumerate}
Note that the reformulation of shared control arbitration (\Cref{sec:problem}) results in the problem of passivating variable impedance, with the solutions presented in \Cref{sec:stabilization} being applicable to any variable impedance control problem.
Table \ref{tab:notation} summarizes the key notations used in our work.

%%%%%%%%%%%%%%%%%%%%%%%%%%%%%%%%%%%%%%%%%%%%%%%%%%%%%%%%%%%%%%%%%%%%%%%%%%%%%%%%
\section{Preliminaries}

\subsection{Passivity}
Passivity of the dynamical system with state model
\begin{align}
	\dot{\bm{x}} &= f(\bm{x}, \bm{u})\\
	\bm{y} &= h(\bm{x}, \bm{u})
\end{align}
is assured if a continuously differentiable positive semidefinite storage function $V(\bm{x})$ exists such that $\bm{u}^\top \bm{y} \ge \dot{V}$ for all $(\bm{x}, \bm{u})$ \cite[Definition 6.3]{khalil2002nonlinear}.
The parallel interconnection of passive systems is also passive, allowing us to sum wrenches of multiple passivated variable impedance controllers.

\subsection{Impedance Control on Riemannian Manifolds}
\label{sec:preliminaries:impedance_control}
We assume a shared control method computing $N$ proxy points $\bm{x}_i$ with associated stiffness $\bm{K}_{\mathrm{p},i}$ and damping $\bm{K}_{\mathrm{d},i}$ matrices.
With a torque controlled manipulator, joint torques $\bm{\tau}$ for weighted Euclidean attractors $\bm{x}_i$ are computed to \cite{hogan1984impedance}
\begin{align}
	\bm{\tau} &= \bm{J}^\top \bm{w}_\mathrm{SC}, \quad \bm{w}_\mathrm{SC} = \sum_{i=1}^N \bm{S}_i \bm{w}_{i,\mathrm{cart}} \label{eq:impedance_to_torque}\\
	\bm{w}_{i,\mathrm{cart}} &=\bm{K}_{\mathrm{p},i} (\bm{x}_i - \bm{x}_\mathrm{ee}) + \bm{K}_{\mathrm{d},i} (\dot{\bm{x}}_i - \dot{\bm{x}}_\mathrm{ee}) \label{eq:individual_wrench}
\end{align}
where $\bm{J}$ denotes the Jacobian of the end effector, $\bm{w}_\mathrm{SC}$ the sum of $\bm{w}_{i,\mathrm{cart}}$ weighted by $\bm{S}_i$ and $\bm{\tau}$ the joint torques.

To also control the orientation, we represent the pose $\bm{x}_i$ using a three-dimensional Euclidean position and a quaternion on the Riemannian manifold $\mathbb{R}^3 \times \mathrm{SO}(3)$.
We treat $\bm{q}$ as the same rotation as $-\bm{q}$.
To take the geometry of the control problem into account, the attractor can also be expressed in cylindrical or spherical coordinates \cite{ti2023geometric} or even on arbitrary meshes \cite{dyck2022impedance}.
The wrench $\bm{w}_i$ \eqref{eq:individual_wrench} then evaluates to \cite{muehlbauer2024probabilistic,muehlbauer2025unified}
\begin{equation}
	\bm{w}_i = \bm{K}_{\mathrm{p},i} \Delta \bm{x}_i + \bm{K}_{\mathrm{d},i} \Delta \dot{\bm{x}}_i\label{eq:impedance_control}
\end{equation}
where $\Delta \bm{x}_i = \mathrm{Log}_{\bm{x}_{\mathrm{ee}}}\left(\bm{x}_i\right)$ denotes the logarithm map\footnote{We wrap $\mathrm{Log}_{\bm{q}}(-\bm{q}) = \bm{0}$ to avoid issues with $\mathcal{S}^3$ double-covering $\mathrm{SO}(3)$.} of $\bm{x}_i$ at $\bm{x}_\mathrm{ee}$, thus computing the geodesic from the end effector towards the attractor point using the default metric of the manifold.
The velocity is given as $\Delta \dot{\bm{x}}_i = \dot{\bm{x}}_i - \dot{\bm{x}}_\mathrm{ee}$, where $\dot{\bm{x}}_i = -\bm{R}_{i,\mathrm{ee}}\mathrm{Log}_{\bm{x}_{i,t}} \left( \bm{x}_{i,t-1} \right) / \Delta t$ with $\bm{R}_{i,\mathrm{ee}} \in \mathbb{R}^{6 \times 6}$ rotating the twist in end effector coordinates and $\dot{\bm{x}}_\mathrm{ee} = \bm{J}_{i,\mathcal{M}} \bm{J} \dot{\bm{q}}$.
The damping matrix $\bm{K}_{\mathrm{d},i}$ is often calculated from the desired stiffness $\bm{K}_{\mathrm{p},i}$ and the mass matrix $\bm{M}$ of the robot using double diagonalization \cite{albuschaeffer2003cartesianimpedance} to achieve an optimally damped behavior.
In the case of $\mathbb{R}^3 \times \mathrm{SO}(3)$, \eqref{eq:impedance_control} leads to a potential similar to \cite{fasse1997spatial,zhang2000spatial} derived for rotation matrices and quaternions.

In case of a manifold other than $\mathbb{R}^3 \times \mathrm{SO}(3)$, the on-manifold wrench is transformed to Cartesian space \cite{dyck2022impedance,muehlbauer2025unified}
\begin{equation}
	\bm{w}_{i,\mathrm{cart}} = \bm{R}_{i,\mathcal{M}} \bm{J}_{i,\mathcal{M}}^T \bm{w}_{i,\mathcal{M}} \label{eq:wrench_manifold_trafo}
\end{equation}
with the rotation matrix $\bm{R}_{i,\mathcal{M}}$ aligning the manifold with the end effector and the Jacobian $\bm{J}_{i,\mathcal{M}}$ transforming the manifold wrench to $\mathbb{R}^3 \times \mathrm{SO}(3)$.
See \cite{muehlbauer2025unified} for details of this operation.
The transformed wrench can then be used in \eqref{eq:impedance_to_torque}.

\begin{table}\caption{\centering \scshape Key notations used in our method.}
	\centering % to have the caption near the table
	\begin{tabular}{l l p{0.55\columnwidth} }
		\toprule
		% -- Robot variables --
		\rowcolor{Light0}
		$\bm{M}$ & $\triangleq$ & Robot mass matrix\\
		\rowcolor{Light0}
		$\bm{x}_\mathrm{ee}$, $\dot{\bm{x}}_\mathrm{ee}$ & $\triangleq$ & Robot end effector pose and velocity\\

		% -- General variables --
		\rowcolor{Light1}
		$\bm{K}_{\mathrm{p},i}$, $\bm{K}_{\mathrm{d},i}$ & $\triangleq$ & Stiffness and damping for attractor $i$\\
		\rowcolor{Light1}
		$\bm{S}_i$ & $\triangleq$ & Scaling matrix for attractor $i$\\
		\rowcolor{Light1}
		$\bm{\mu}_i$, $\bm{\Sigma}_i$ & $\triangleq$ & Mean and covariance for attractor $i$\\
		\rowcolor{Light1}
		$\bm{w}_i$ & $\triangleq$ & Wrench for attractor $i$\\
		\rowcolor{Light1}
		$\bm{x}_i$ & $\triangleq$ & Pose of attractor $i$\\

		% -- transformations --
		\rowcolor{Light2}
		$\bm{R}_{i,\mathcal{M}}$ & $\triangleq$ & Coordinate rotation of attractor $i$ \\
		\rowcolor{Light2}
		$\bm{J}_{i,\mathcal{M}}$ & $\triangleq$ & Coordinate Jacobian of attractor $i$\\

		% -- general values of the method --
		\rowcolor{Light3}
		$\bm{K}^*_{\mathrm{p},i}$ & $\triangleq$ & Product of scalings $\bm{S}_i$ and stiffness $\bm{K}_{\mathrm{p},i}$\\
		\rowcolor{Light3}
		$V_{\mathrm{ag},i}$ & $\triangleq$ & Storage function of attractor $i$ \\
		\rowcolor{Light3}
		$\Delta \bm{x}_i = \mathrm{Log}_{\bm{x}_{\mathrm{ee}}}\left(\bm{x}_i\right)$ & $\triangleq$ & Spring deflection of attractor $i$ \\
		\rowcolor{Light3}
		$\Delta \dot{\bm{x}}_i = \dot{\bm{x}}_i - \dot{\bm{x}}_\mathrm{ee}$ & $\triangleq$ & Spring velocity of attractor $i$ \\
		\rowcolor{Light3}
		$d_i$, $\dot{d}_i$ & $\triangleq$ & Scaling and its derivative for attractor $i$ \\

		% -- passivations --
		\rowcolor{Light4}
		$d_i \mathrm{Log}_{\bm{x}_{\mathrm{ee}}}\left(\bm{x}_i\right)$ & $\triangleq$ & Passivated deflection (\Cref{sec:stabilization:spring_deflection}) \\
		\rowcolor{Light4}
		$\bm{K}^+_{\mathrm{p},i}$ & $\triangleq$ & Passivated stiffness (\Cref{sec:stabilization:stiffness_change}) \\
		\bottomrule
	\end{tabular}
	\vspace{-1em}
	\label{tab:notation}
\end{table}

\subsection{Adaptive Shared Control}
\label{sec:preliminaries:shared_control}
Advanced shared control approaches use both variable impedance behaviours \textit{within} constraints \cite{muehlbauer2025unified,michel2023learning} as well as arbitration schemes \textit{combining} wrenches $\bm{w}_i$ of constraints~\cite{raiola2017comanipulation,selvaggio2016enhancing,balachandran2022stable,balachandran2023passive,muehlbauer2025unified}.
Passivity of the arbitration $\bm{w} = \sum_i^N \alpha_i \bm{w}_i$, though at the cost of position drift, using factors $\alpha_i$ with $\sum_{i=1}^N \alpha_i = 1$ has already been explored in \cite{balachandran2022stable,balachandran2023passive}.

An optimization-based approach to the arbitration of probabilistic wrenches $\bm{w}_i \sim \mathcal{N}(\bm{\mu}_i, \bm{\Sigma}_i)$ as proposed in \cite{muehlbauer2025unified}
\begin{equation}
	\hat{\bm{w}} = \mathrm{arg}~\underset{\bm{w}}{\mathrm{min}} \sum_{i=1}^N \left( \bm{w} - \bm{\mu}_i \right)^\top \bm{\Sigma}^{-1}_i \left( \bm{w} - \bm{\mu}_i \right),\label{eq:poe_wrench_optimization}
\end{equation}
is solved as product of $N$ Gaussians
\begin{equation}
	\hat{\bm{w}} = \hat{\bm{\Sigma}} \sum_{i=1}^N \bm{\Sigma}^{-1}_i \bm{\mu}_i, \quad \hat{\bm{\Sigma}} = \left( \sum_{i=1}^N \bm{\Sigma}^{-1}_i \right)^{-1}.
	\label{eq:vf_covariance_arbitration}
\end{equation}
This approach leads to matrix-valued scaling factors $\bm{S}_i = \hat{\bm{\Sigma}} \bm{\Sigma}^{-1}_i$ in the arbitration equation \eqref{eq:impedance_to_torque} requiring a more extensive treatment of passivity.
Note that $\bm{S}_i$  is not necessarily symmetric.
In this work, we provide methods for stabilizing both time-varying stiffness matrices $\bm{K}_{\mathrm{p},i}$ \eqref{eq:impedance_control} as well as time-varying arbitration $\bm{S}_i$ \eqref{eq:impedance_to_torque}.

%%%%%%%%%%%%%%%%%%%%%%%%%%%%%%%%%%%%%%%%%%%%%%%%%%%%%%%%%%%%%%%%%%%%%%%%%%%%%%%%
\section{Problem Analysis}
\label{sec:problem}
\begin{figure}
	\centering
	\tikzstyle{l} = [draw, -latex',thick]
	\resizebox{0.8\columnwidth}{!}{%
	\begin{tikzpicture}
		% System border
		\node[rectangle,draw=black,dashed,inner sep=0pt,minimum width=12.5cm,minimum height=11cm] at (1cm, 0cm) {};

		% Scaling
		\node[rectangle,draw=black,inner sep=0pt,minimum width=2cm,minimum height=10cm] at (0, 0) (scaling) {Scaling};

		% Controller 1
		\node[rectangle,draw=black,inner sep=0pt,minimum width=2cm,minimum height=2.5cm,above left=-2.5cm and 2cm of scaling] (ctrl1) {Ctrl. 1};
		\node[minimum width=2cm,minimum height=2.5cm,above left=-2.5cm and 0cm of scaling] {$\bm{w}_1$};
		\node[minimum width=2cm,minimum height=2.5cm,left=-0.3cm of ctrl1] {$\bm{w}_1$};
		\draw ([yshift=0.8cm]ctrl1.east)  -- ([yshift=4.55cm]scaling.west) node [midway, below] {$+$} node [midway, above,align=center] {$\dot{\bm{x}}_\mathrm{ee}$ \\ $\rightarrow$};
		\draw ([yshift=-0.8cm]ctrl1.east)  -- ([yshift=2.95cm]scaling.west) node [midway, above] {$-$};
		\node[circle,fill=black,inner sep=0pt,minimum size=5pt,above left=-0.52cm and 1cm of ctrl1] (ctlr1_in1) {};
		\node[circle,fill=black,inner sep=0pt,minimum size=5pt,below left=-0.52cm and 1cm of ctrl1] (ctlr1_in2) {};
		\draw ([yshift=0.8cm]ctrl1.west)  -- (ctlr1_in1) node [near end, below] {$+$} node [near end, above,align=center] {$\dot{\bm{x}}_1$ \\ $\rightarrow$};
		\draw ([yshift=-0.8cm]ctrl1.west)  -- (ctlr1_in2) node [near end, above] {$-$};

		% Controller 2
		\node[rectangle,draw=black,inner sep=0pt,minimum width=2cm,minimum height=2.5cm,above left=-5.5cm and 2cm of scaling] (ctrl2) {Ctrl. 2};
		\node[minimum width=2cm,minimum height=2.5cm,above left=-5.5cm and 0cm of scaling] {$\bm{w}_2$};
		\node[minimum width=2cm,minimum height=2.5cm,left=-0.3cm of ctrl2] {$\bm{w}_2$};
		\draw ([yshift=0.8cm]ctrl2.east)  -- ([yshift=1.55cm]scaling.west) node [midway, below] {$+$} node [midway, above,align=center] {$\dot{\bm{x}}_\mathrm{ee}$ \\ $\rightarrow$};
		\draw ([yshift=-0.8cm]ctrl2.east)  -- ([yshift=-0.05cm]scaling.west) node [midway, above] {$-$};
		\node[circle,fill=black,inner sep=0pt,minimum size=5pt,above left=-0.52cm and 1cm of ctrl2] (ctlr2_in1) {};
		\node[circle,fill=black,inner sep=0pt,minimum size=5pt,below left=-0.52cm and 1cm of ctrl2] (ctlr2_in2) {};
		\draw ([yshift=0.8cm]ctrl2.west)  -- (ctlr2_in1) node [near end, below] {$+$} node [near end, above,align=center] {$\dot{\bm{x}}_2$ \\ $\rightarrow$};
		\draw ([yshift=-0.8cm]ctrl2.west)  -- (ctlr2_in2) node [near end, above] {$-$};
		% Dots
		\node[inner sep=0pt,minimum width=2cm,minimum height=2.5cm,above left=-7.75cm and 2cm of scaling] (ctrldots) {\vdots};

		% Controller n
		\node[rectangle,draw=black,inner sep=0pt,minimum width=2cm,minimum height=2.5cm,above left=-10cm and 2cm of scaling] (ctrln) {Ctrl. n};
		\node[minimum width=2cm,minimum height=2.5cm,above left=-10cm and 0cm of scaling] {$\bm{w}_n$};
		\node[minimum width=2cm,minimum height=2.5cm,left=-0.3cm of ctrln] {$\bm{w}_n$};
		\draw ([yshift=0.8cm]ctrln.east)  -- ([yshift=-2.95cm]scaling.west) node [midway, below] {$+$} node [midway, above,align=center] {$\dot{\bm{x}}_\mathrm{ee}$ \\ $\rightarrow$};
		\draw ([yshift=-0.8cm]ctrln.east)  -- ([yshift=-4.55cm]scaling.west) node [midway, above] {$-$};
		\node[circle,fill=black,inner sep=0pt,minimum size=5pt,above left=-0.52cm and 1cm of ctrln] (ctlrn_in1) {};
		\node[circle,fill=black,inner sep=0pt,minimum size=5pt,below left=-0.52cm and 1cm of ctrln] (ctlrn_in2) {};
		\draw ([yshift=0.8cm]ctrln.west)  -- (ctlrn_in1) node [near end, below] {$+$} node [near end, above,align=center] {$\dot{\bm{x}}_n$ \\ $\rightarrow$};
		\draw ([yshift=-0.8cm]ctrln.west)  -- (ctlrn_in2) node [near end, above] {$-$};

		% Robot
		\node[rectangle,draw=black,inner sep=0pt,minimum width=2cm,minimum height=2.5cm,right=4cm of scaling] (robot) {Robot};
		\node[minimum width=2cm,minimum height=2.5cm,right=0.5cm of scaling] {$\bm{S}_1 \bm{w}_1 + \dots + \bm{S}_n \bm{w}_n$};
		\draw ([yshift=0.8cm]scaling.east)  -- ([yshift=0.8cm]robot.west) node [midway, below] {$+$} node [midway, above,align=center] {$\dot{\bm{x}}_\mathrm{ee}$ \\ $\rightarrow$};
		\draw ([yshift=-0.8cm]scaling.east)  -- ([yshift=-0.8cm]robot.west) node [midway, above] {$-$};
		\node[circle,fill=black,inner sep=0pt,minimum size=5pt,above right=-0.52cm and 1cm of robot] (robot_in1) {};
		\node[circle,fill=black,inner sep=0pt,minimum size=5pt,below right=-0.52cm and 1cm of robot] (robot_in2) {};
		\draw ([yshift=0.8cm]robot.east)  -- (robot_in1) node [near end, below] {$+$} node [near end, above,align=center] {$\dot{\bm{x}}_\mathrm{ee}$ \\ $\rightarrow$};
		\draw ([yshift=-0.8cm]robot.east)  -- (robot_in2) node [near end, above] {$-$};
		\node[minimum width=2cm,minimum height=2.5cm,right=-0.3cm of robot] {$\bm{w}_e$};
	\end{tikzpicture}}
	\caption{\label{fig:port_network_overview}Port-network representation of the shared control system with $n$ controllers acting on the robot, with $3$ controllers shown in the image.\vspace{-1em}}
\end{figure}
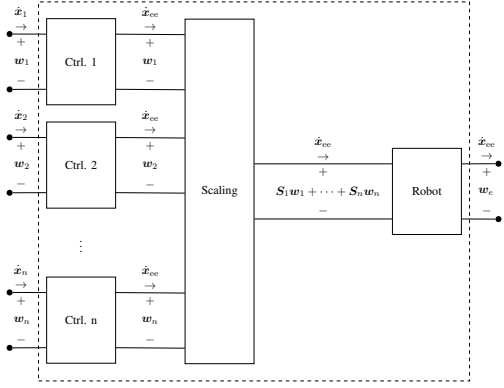
The shared control system of \Cref{sec:preliminaries:shared_control} is visualized as port-network diagram in \Cref{fig:port_network_overview}.
We ensure passivity with respect to the power ports shown therein.
Scaling of the controller wrenches $\bm{S}_i \bm{w}_i$ by $\bm{S}_i$ is reformulated as stiffness scaling in \textit{Ctrl. i} which we then passivate.
Our methods derived in \Cref{sec:stabilization} hold for variable impedance control in general.

\subsection{Arbitration as Stiffness Modulation}
\label{sec:arbitration_as_modulation}
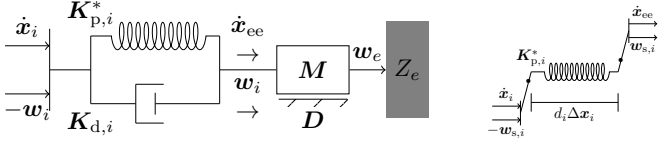
\begin{figure}
	\centering
	\tikzstyle{l} = [draw, -latex',thick]
\begin{subfigure}{0.65\columnwidth}
	\resizebox{\columnwidth}{!}{%
	\begin{tikzpicture}
		% Ground
		\draw [pattern={Lines[angle=45,distance=7pt]},pattern color=black,draw=none] (0,0) rectangle (1,-0.1) node[midway,below] {$\bm{D}$};
		\draw (0,0) -- ++ (1,0) node [midway, above=2pt, rectangle, draw, minimum width=30pt, minimum height=20pt] (mass) {$\bm{M}$};

		% External wrench
		\node[rectangle,fill=gray,minimum width=15pt, minimum height=40pt,right=15pt of mass] (fe) {$Z_e$};
		\draw[<-] (fe.west) -- (mass.east) node[midway, above] {$\bm{w}_e$};

		% Spring and damper support
		\node[left=20pt of mass] (springdamper_right) {};
		\node[left=50pt of springdamper_right.center] (springdamper_left) {};
		\draw (mass.west) -- (springdamper_right.center) node[midway, above,align=center] {$\dot{\bm{x}}_\mathrm{ee}$\\$\rightarrow$} node[midway, below,align=center] {$\bm{w}_i$\\$\rightarrow$};

		% Spring
		\node[above=7pt of springdamper_right] (spring_right) {};
		\node[above=7pt of springdamper_left] (spring_left) {};
		\draw (spring_right.center) -- (springdamper_right.center);
		\draw (spring_left.center) -- (springdamper_left.center);
		\draw[decoration={
			coil,
			aspect=0.3,
			segment length=1.2mm,
			amplitude=2mm,
			pre length=2mm,
			post length=2mm},
			decorate] (spring_right.center) -- (spring_left.center) node[above,black]{$\bm{K}^*_{\mathrm{p},i}$};

		% Damper
		\node[below=7pt of springdamper_right] (damper_right) {};
		\node[below=7pt of springdamper_left] (damper_left) {};
		\draw (damper_right.center) -- (springdamper_right.center);
		\draw (damper_left.center) -- (springdamper_left.center);
		\node (damper_center) at ($(damper_right)!0.5!(damper_left)$) {};
		\node[above=1pt of damper_center] (damper_center_above) {};
		\node[below=1pt of damper_center] (damper_center_below) {};
		\draw (damper_right.center) -- (damper_center.center);
		\draw (damper_center_above.south) -- (damper_center.center);
		\draw (damper_center_below.north) -- (damper_center.center);
		\node[left=1pt of damper_center] (damper_center_left) {};
		\node[above=1pt of damper_center_left] (damper_left_center_above) {};
		\node[below=1pt of damper_center_left] (damper_left_center_below) {};
		\draw (damper_center_left.center) -- (damper_left.center) node[below,black]{$\bm{K}_{\mathrm{d},i}$};
		\draw (damper_left_center_above.center) -- (damper_center_left.center);
		\draw (damper_left_center_below.center) -- (damper_center_left.center);
		\draw (damper_center_above.east) -- (damper_left_center_above.center);
		\draw (damper_center_below.east) -- (damper_left_center_below.center);

		% Left support
		\node[left=10pt of springdamper_left] (leftsupport) {};
		\node[above=10pt of leftsupport.center] (above_leftsupport) {};
		\node[below=10pt of leftsupport.center] (below_leftsupport) {};
		\draw (springdamper_left.center) -- (leftsupport.center);
		\draw (leftsupport.center) -- (above_leftsupport.north);
		\draw (leftsupport.center) -- (below_leftsupport.south);
		
		% Left stuff
		\node[left=15pt of above_leftsupport.south] (arrow_above_leftsupport) {};
		\node[left=15pt of below_leftsupport.north] (arrow_below_leftsupport) {};
		\draw[->] (arrow_above_leftsupport.center) -- (above_leftsupport.south) node[midway, above] {$\dot{\bm{x}}_i$};
		\draw[->] (arrow_below_leftsupport.center) -- (below_leftsupport.north) node[midway, below] {$-\bm{w}_i$};
	\end{tikzpicture}}
\end{subfigure}
\hfill
\begin{subfigure}{0.27\columnwidth}
	\resizebox{\columnwidth}{!}{%
	\begin{tikzpicture}
		% Spring
		\node at (0,0) (spring_right) {};
		\node[left=50pt of spring_right] (spring_left) {};
		\draw[decoration={
			coil,
			aspect=0.3,
			segment length=1.2mm,
			amplitude=2mm,
			pre length=2mm,
			post length=2mm},
			decorate] (spring_right.center) -- (spring_left.center) node[above,black]{$\bm{K}^*_{\mathrm{p},i}$};

		% spring length
		\node[below=7pt of spring_right] (right_dist_up) {};
		\node[below=5pt of right_dist_up] (right_dist_down) {};
		\draw (right_dist_up.center) -- (right_dist_down.center) node[midway] (right_dist_center) {};
		\node[below=7pt of spring_left] (left_dist_up) {};
		\node[below=5pt of left_dist_up] (left_dist_down) {};
		\draw (left_dist_up.center) -- (left_dist_down.center) node[midway] (left_dist_center) {};
		\draw (right_dist_center.center) -- (left_dist_center.center) node [midway, below] {$d_i \Delta \bm{x}_i$};

		% right lever
		\node[above right=20pt and 0pt of spring_right] (right_lever) {};
		\draw (spring_right.center) -- (right_lever.center) node[near start]{\textbullet};
		\node[above=1pt of right_lever] (right_lever_above) {};
		\node[below=1pt of right_lever] (right_lever_below) {};
		\draw (right_lever_above.center) -- (right_lever_below.center);
		\node[right=15pt of right_lever_above.south] (arrow_above_rightsupport) {};
		\node[right=15pt of right_lever_below.north] (arrow_below_rightsupport) {};
		\draw[<-] (arrow_above_rightsupport.center) -- (right_lever_above.south) node[midway, above] {$\dot{\bm{x}}_\mathrm{ee}$};
		\draw[<-] (arrow_below_rightsupport.center) -- (right_lever_below.north) node[midway, below] {$\bm{w}_{\mathrm{s},i}$};

		% left lever
		\node[below left=20pt and 0pt of spring_left] (left_lever) {};
		\draw (spring_left.center) -- (left_lever.center) node[near start]{\textbullet};
		\node[above=1pt of left_lever] (left_lever_above) {};
		\node[below=1pt of left_lever] (left_lever_below) {};
		\draw (left_lever_above.center) -- (left_lever_below.center);
		\node[left=15pt of left_lever_above.south] (arrow_above_leftsupport) {};
		\node[left=15pt of left_lever_below.north] (arrow_below_leftsupport) {};
		\draw[->] (arrow_above_leftsupport.center) -- (left_lever_above.south) node[midway, above] {$\dot{\bm{x}}_i$};
		\draw[->] (arrow_below_leftsupport.center) -- (left_lever_below.north) node[midway, below] {$-\bm{w}_{\mathrm{s},i}$};
	\end{tikzpicture}}
\end{subfigure}
	\caption{\label{fig:kp_mod_mechanics}Mechanical equivalent of the modified stiffness gains (left) and lever-based scaling of the spring (right). Through the lever support (black dot), $\dot{\bm{x}}_i$ and $\dot{\bm{x}}_\mathrm{ee}$ are scaled by $d$ towards the spring whereas the spring wrenches $\bm{w}_{\mathrm{s},i}$ are the wrenches generated by the spring also scaled with $d$.\vspace{-1em}}
\end{figure}
Assuming access to the impedance controller's stiffness matrices, we can reformulate the arbitration from~\eqref{eq:impedance_to_torque} into a stiffness variation.
For Cartesian controllers, a modified stiffness matrix $\bm{K}'_{\mathrm{p},i}$ can be computed as $\bm{K}'_{\mathrm{p},i} = \bm{S}_i \bm{K}_{\mathrm{p},i}$.
For transformed manifold-valued stiffnesses \eqref{eq:wrench_manifold_trafo}, we get
\begin{equation}
	\bm{S}_i \bm{R}_{i,\mathcal{M}} \bm{J}_{i,\mathcal{M}}^T \bm{K}_{\mathrm{p},i} = \bm{R}_{i,\mathcal{M}} \bm{J}_{i,\mathcal{M}}^T \bm{K}'_{\mathrm{p},i}.
\end{equation}
Assuming the existence of $(\bm{J}_{i,\mathcal{M}}^T)^{-1}$ which holds away from singularities, we obtain
\begin{equation}
	\label{eq:arbitration_to_stiffness_variation}
	\bm{K}'_{\mathrm{p},i} = (\bm{J}_{i,\mathcal{M}}^T)^{-1} \bm{R}_{i,\mathcal{M}}^{-1} \bm{S}_i \bm{R}_{i,\mathcal{M}} \bm{J}_{i,\mathcal{M}}^T \bm{K}_{\mathrm{p},i}.
\end{equation}
We subsequently apply symmetrization to $\bm{K}'_{\mathrm{p},i}$ to obtain symmetric ($\bm{K}^*_{\mathrm{p},i}$) and asymmetric ($\bm{K}^{*}_{\mathrm{p},\mathrm{curl},i})$ components
\begin{gather}
	\bm{K}^*_{\mathrm{p},i} = \frac{1}{2} \left(\bm{K}'_{\mathrm{p},i} + {\bm{K}'_{\mathrm{p},i}}^\top \right), \quad \bm{K}^{*}_{\mathrm{p},\mathrm{curl},i} = \frac{1}{2} \left(\bm{K}'_{\mathrm{p},i} - {\bm{K}'_{\mathrm{p},i}}^\top \right).\nonumber
\end{gather}

A mechanical equivalent of such system with only symmetric modified stiffness matrices $\bm{K}^*_{\mathrm{p},i}$ is drawn in \Cref{fig:kp_mod_mechanics}.

At any given time, $N$ of these springs act concurrently on the robot's end effector.
As the parallel interconnection of passive systems remains passive, it is sufficient to ensure the passivity in each spring-damper system separately.

\subsection{Passivity of Variable-Impedance Controllers}
In case of symmetric constant stiffness springs, the system shown in \Cref{fig:kp_mod_mechanics} would be passive.
The storage function
\begin{equation}
	V = \frac{1}{2} \dot{\bm{x}}_\mathrm{ee}^\top \bm{M} \dot{\bm{x}}_\mathrm{ee} + \sum_i \frac{1}{2} \Delta \bm{x}_i^\top \bm{K}^*_{\mathrm{p},i} \Delta \bm{x}_i
	\label{eq:storage_function_initial}
\end{equation}
combining both kinetic energy related to the robot inertia and velocity and potential energy associated with the springs, reveals that an increasing stiffness matrix $\bm{K}^*_{\mathrm{p},i}$ may lead to a non-passive system \cite{balachandran2022stable,balachandran2023passive} as the energy stored in the springs increases.
The passivity violating power of this stiffness increase must be compensated for to ensure passivity in the system.
Power is then only introduced through the power ports, e.g. via an external force from a human operator, which can then be used to increase the stiffness.
To ensure passivity, the rate of energy increase in the spring's storage function has to be limited to the available dissipating terms.
This can be achieved by either limiting the rate of stiffness increase $\dot{\bm{K}}^*_{\mathrm{p},i}$ or by limiting the spring deflection $\Delta \bm{x}_i = \mathrm{Log}_{\bm{x}_{\mathrm{ee}}}\left(\bm{x}_i\right)$.
We will explore both methods in the following.

%%%%%%%%%%%%%%%%%%%%%%%%%%%%%%%%%%%%%%%%%%%%%%%%%%%%%%%%%%%%%%%%%%%%%%%%%%%%%%%%
\section{Proposed Passivation Methods}
\label{sec:stabilization}
We first provide passivation approaches for the symmetrized $\bm{K}^*_{\mathrm{p},i}$ in \Cref{sec:stabilization:spring_deflection,sec:stabilization:stiffness_change}.
We then show the passivation of the asymmetric component $\bm{K}^{*}_{\mathrm{p},\mathrm{curl},i}$ (\Cref{sec:stabilization:extensions}).

\subsection{Limiting the Spring Deflection}
\label{sec:stabilization:spring_deflection}
As first approach, we propose to limit the spring deflection using a scalar factor $0 \le d_i \le 1$ to $d_i \Delta \bm{x}_i$.
The scaling factor $d_i = 0$ corresponds to a completely disabled spring while $d_i = 1$ corresponds to the nominal spring behaviour as requested by the shared control system.
Values $0 < d_i < 1$ scale the spring.
This leads to a modified mechanical system (\Cref{fig:kp_mod_mechanics}), where the spring is connected to the robot $\bm{x}_\mathrm{ee}$ as well as to the $i$-th attractor $\bm{x}_i$ through a lossless lever with transmission ratio $d_i$.
The spring is therefore only deflected by $d_i \Delta \bm{x}_i$, and the velocities entering the spring from either side are also scaled by a factor of $d_i$.
Furthermore, the wrench $\bm{w}_{\mathrm{s},i}$ acting on the robot respectively the environment is scaled by $d_i$ to preserve power flowing over the lever as given by
\begin{align}
	P_{\mathrm{s},i} &= \left( \bm{K}_{\mathrm{p},i}^* d_i \Delta \bm{x}_i \right)^\top d_i \Delta \dot{\bm{x}}_i \overset{!}{=} \bm{w}_{\mathrm{s},i}^\top \Delta \dot{\bm{x}}_i,
\end{align}
resulting in $\bm{w}_i = \bm{w}_{\mathrm{s},i} + \bm{w}_{\mathrm{d},i} = d_i^2 \bm{K}_{\mathrm{p},i}^* \Delta \bm{x}_i + \bm{K}_{\mathrm{d},i} \Delta\dot{\bm{x}}_i$ with damping wrench $\bm{w}_{\mathrm{d},i}$.
The storage function evaluates to
\begin{equation}
	V = \frac{1}{2} \dot{\bm{x}}_\mathrm{ee}^\top \bm{M} \dot{\bm{x}}_\mathrm{ee} + \sum_i \frac{1}{2} d_i^2 \Delta \bm{x}_i^\top \bm{K}^*_{\mathrm{p},i} \Delta \bm{x}_i.\label{eq:storage_fcn}
\end{equation}
Considering $\dot{\bm{x}}_\mathrm{ee}^\top \left(\frac{1}{2} \dot{\bm{M}} - \bm{C}(\bm{q}, \dot{\bm{q}}) \right) \dot{\bm{x}}_\mathrm{ee} = 0$ due to skew symmetry \cite{featherstone2008dynamics}, the derivative of this storage function is
\begin{align}
	\dot{V}& = \dot{\bm{x}}_\mathrm{ee}^\top \bm{M} \ddot{\bm{x}}_\mathrm{ee} + \sum_i d_i \dot{d}_i \Delta\bm{x}_i^\top \bm{K}^*_{\mathrm{p},i} \Delta\bm{x}_i\nonumber\\
	&+ \sum_i d_i^2 \Delta\bm{x}_i^\top \bm{K}^*_{\mathrm{p},i} \Delta\dot{\bm{x}}_i + \sum_i \frac{1}{2} d_i^2 \Delta\bm{x}_i^\top \dot{\bm{K}}^*_{\mathrm{p},i} \Delta\bm{x}_i,\label{eq:power_fcn}
\end{align}
%with assumption of constant $\bm{M}$ (see \Cref{appendix:proof} for a proof that the result does not change with varying $\bm{M}(\bm{q})$)
where
\begin{equation}
	\dot{V} \le \sum_i \dot{\bm{x}}_i^T \bm{w}_i + \dot{\bm{x}}_\mathrm{ee}^T \bm{w}_\mathrm{ee} = \dot{V}_\mathrm{inp}\label{eq:passivity_condition}
\end{equation}
ensures passivity w.r.t. $(\dot{\bm{x}}_i, \bm{w}_i)$ and $(\dot{\bm{x}}_\mathrm{ee}, \bm{w}_\mathrm{ee})$ shown in \Cref{fig:port_network_overview}.
The robot acceleration is caused by wrenches
\begin{align}
	\bm{M} \ddot{\bm{x}}_\mathrm{ee} &= \sum_{i=1}^N \bm{w}_i + \bm{w}_e - \bm{D} \dot{\bm{x}}_\mathrm{ee}, \label{eq:robot_motion}
\end{align}
where $\bm{D}$ represents the combination of physical and low-level controller damping.
Deriving the velocity of the end effector using an inertial frame $\bm{x}_\mathrm{world}$ to $\dot{\bm{x}}_\mathrm{ee} = -\frac{\mathrm{d}}{\mathrm{d}t}\mathrm{Log}_{\bm{x}_{\mathrm{ee}}}\left(\bm{x}_\mathrm{world}\right)$ allows for calculating the velocity of the $i$-th attractor point with respect to an inertial frame in the end effector coordinate system to $\dot{\bm{x}}_i = \Delta \dot{\bm{x}}_i + \dot{\bm{x}}_\mathrm{ee}$, resulting in the identity
\begin{equation}
	\dot{\bm{x}}_i^\top \bm{w}_i = d_i^2 \Delta\dot{\bm{x}}_i^\top \bm{K}^*_{\mathrm{p},i} \Delta \bm{x}_i + \Delta \dot{\bm{x}}_i^\top \bm{K}_{\mathrm{d},i} \Delta \dot{\bm{x}}_i + \dot{\bm{x}}_\mathrm{ee}^\top \bm{w}_i.\label{eq:ext_wrench_identity}
\end{equation}
Substituting $\sum_i \dot{\bm{x}}_\mathrm{ee}^\top \bm{w}_i$ after inserting \eqref{eq:robot_motion} in \eqref{eq:power_fcn} leads to
\begin{align}
	\dot{V} &= \sum_i \dot{\bm{x}}_i^\top \bm{w}_i + \dot{\bm{x}}_\mathrm{ee}^\top \bm{w}_e - \dot{\bm{x}}_\mathrm{ee}^\top \bm{D} \dot{\bm{x}}_\mathrm{ee} - \sum_i \Delta \dot{\bm{x}}_i^\top \bm{K}_{\mathrm{d},i} \Delta \dot{\bm{x}}_i\nonumber\\
	&+ \sum_i d_i \dot{d}_i \Delta\bm{x}_i^\top \bm{K}^*_{\mathrm{p},i} \Delta\bm{x}_i + \sum_i \frac{1}{2} d_i^2 \Delta\bm{x}_i^\top \dot{\bm{K}}^*_{\mathrm{p},i} \Delta\bm{x}_i.\label{eq:power_fcn_subst}
	\raisetag{2em}
\end{align}
Clearly, only the last two terms can violate the passivity condition \eqref{eq:passivity_condition}.
Note that for $d_i = 1$ (with accordingly limited $\dot{d}_i \le 0$), energy can only be generated for the last term with increasing stiffness which is in line with the results of \cite{balachandran2022stable,balachandran2023passive}.
As in their work, we can calculate the damping power
\begin{equation}
	P_{\mathrm{d}, i} = \Delta \dot{\bm{x}}_i^\top \bm{K}_{\mathrm{d},i} \Delta \dot{\bm{x}}_i\label{eq:damping_power}
\end{equation}
which we use to calculate the scaling $d_i$ by ensuring that the possibly passivity-violating terms of every spring are smaller than the corresponding damping power for each controller $i$
\begin{align}
	\underbrace{P_{\mathrm{d},i}}_{\text{damping power}} &\ge \underbrace{d_i \dot{d}_i \Delta\bm{x}_i^\top \bm{K}^*_{\mathrm{p},i} \Delta\bm{x}_i}_{\text{power from $\dot{d}_i$}} + \underbrace{\frac{1}{2} d_i^2 \Delta\bm{x}_i^\top \dot{\bm{K}}^*_{\mathrm{p},i} \Delta\bm{x}_i}_{\text{power from $\dot{\bm{K}}^*_{\mathrm{p},i}$}}.\label{eq:passivity_condition_l}
	\raisetag{0.7em}
\end{align}
The evolution of $d_i$ is thus governed by
\begin{align}
	\dot{d}_i &= \frac{P_{\mathrm{d}, i} - \frac{1}{2} d_i^2 \Delta\bm{x}_i^\top \dot{\bm{K}}^*_{\mathrm{p},i} \Delta\bm{x}_i}{d_i \Delta\bm{x}_i^\top \bm{K}^*_{\mathrm{p},i} \Delta\bm{x}_i + \epsilon}. \label{eq:ddot_evolution}
\end{align}
%which can be computed and subsequently integrated inside the controller.
A small $\epsilon > 0$ avoids division by zero when $d_i \Delta\bm{x}_i^\top \bm{K}^*_{\mathrm{p},i} \Delta\bm{x}_i = 0$; this is especially important to allow the scaling factor to increase again after becoming $0$ or when the control error $\Delta \bm{x}_i$ vanishes.
We obtain the desired scaling $d_i$ by integrating $\dot{d}_i$ and limiting it to $0 \le d_i \le 1$.

\begin{figure}
	\centering
    \includegraphics[width=\columnwidth,trim={30, 20, 20, 15},clip]{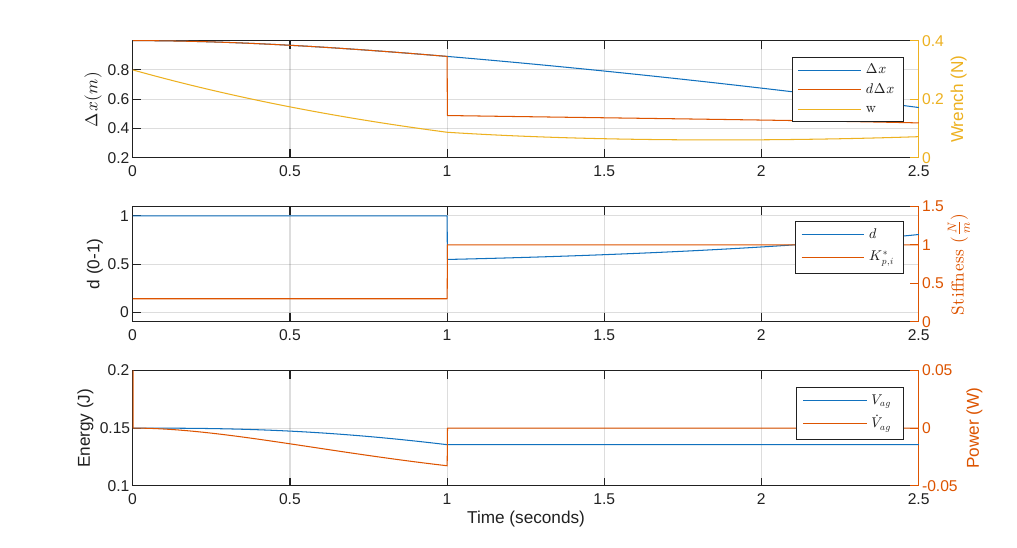}
	\caption{\label{fig:limit_spring_deflection} Limiting the deflection $\Delta \bm{x}$ to $d \Delta \bm{x}$ in continuous time simulation through \eqref{eq:impedance_control_pass_l}. At $t=1$, we simulate a step in the desired stiffness $\bm{K}_{\mathrm{p},i}^*$.\vspace{-1em}}
\end{figure}

We can now update the impedance control law \eqref{eq:impedance_control} to
\begin{equation}
	\bm{w}_{\mathrm{pass},i} = \bm{K}^*_{\mathrm{p},i} d_i^2 \mathrm{Log}_{\bm{x}_{\mathrm{ee}}}\left(\bm{x}_i\right) + \bm{K}_{\mathrm{d},i} \Delta \dot{\bm{x}}_i,\label{eq:impedance_control_pass_l}
\end{equation}
resulting in a passivated wrench $\bm{w}_{\mathrm{pass},i}$.
\Cref{fig:limit_spring_deflection} shows the behavior of this approach with a simulated 1-\gls{dof} system to a step in the desired stiffness.
As \eqref{eq:ddot_evolution} is a very stiff equation for stiffness steps, implementation in a time-discrete controller is challenging.
In order to support arbitrary stiffness changes, a discrete time treatment of the problem is necessary.

To this end, \eqref{eq:storage_fcn}, \eqref{eq:robot_motion} and \eqref{eq:ext_wrench_identity} can be reformulated for a discretized system.
%Assuming constant wrenches and velocities for one time step $\Delta t$, we calculate the energy difference to
% and neglecting the position change between two time steps, we can calculate the energy difference between two time steps based on \eqref{eq:power_fcn_subst}
We calculate the energy difference to
\begin{align}
	V_\mathrm{k+1} - V_\mathrm{k} &= \left(\sum_i \dot{\bm{x}}_{i,k}^\top \bm{w}_{i,k} - \sum_i P_{\mathrm{d},i,k}\right.&\tikzmark{ekin_start}\nonumber\\
	&~~~+ \dot{\bm{x}}_{\mathrm{ee},k}^\top \bm{w}_{e,k} - \dot{\bm{x}}_{\mathrm{ee},k}^\top \bm{D} \dot{\bm{x}}_{\mathrm{ee},k}&\nonumber\\
	&~~~\left.- \sum_i d_{i,k}^2 \Delta \dot{\bm{x}}_{i,k}^\top \bm{K}_{\mathrm{p},i,k}^* \Delta \bm{x}_{i,k-1}\right) \Delta t&\tikzmark{ekin_end}\nonumber\\
	&~~~+ \sum_i \frac{1}{2} d_{i,k+1}^2 \Delta\bm{x}_{i,k}^\top \bm{K}_{\mathrm{p},i,k+1}^* \Delta\bm{x}_{i,k}&\tikzmark{epot_start}\nonumber\\
	&~~~- \sum_i \frac{1}{2} d_{i,k}^2 \Delta\bm{x}_{i,k-1}^\top \bm{K}_{\mathrm{p},i,k}^* \Delta\bm{x}_{i,k-1}.&\tikzmark{epot_end}\nonumber
\end{align}
For $\frac{1}{2} \Delta\bm{x}_{i,k-1}^\top \bm{K}_{\mathrm{p},i,k}^* \Delta\bm{x}_{i,k-1} + \Delta \dot{\bm{x}}_{i,k}^\top \bm{K}_{\mathrm{p},i,k}^* \Delta \bm{x}_{i,k-1} \Delta t \approx \frac{1}{2} \Delta\bm{x}_{i,k}^\top \bm{K}_{\mathrm{p},i,k}^* \Delta\bm{x}_{i,k}$, passivity \eqref{eq:passivity_condition} is given through the approximation (violations near numerical noise, \Cref{sec:eval:rotations_impacts})
\begin{align}
	\underbrace{P_{\mathrm{d},i,k} \Delta t}_{\text{damped energy}} + \underbrace{\frac{1}{2} d_{i,k}^2 \Delta \bm{x}_{i,k}^\top \bm{K}_{\mathrm{p},i,k}^* \Delta \bm{x}_{i,k}}_{\text{spring energy with $\bm{K}_{\mathrm{p},i,k}$}}\nonumber\\
	\ge \underbrace{\frac{1}{2} d_{i,k}^2 \Delta \bm{x}_{i,k}^\top \bm{K}_{\mathrm{p},i,k+1}^* \Delta \bm{x}_{i,k}}_{\text{spring energy with $\bm{K}_{\mathrm{p},i,k+1}$}}.\label{eq:spring_deflection:discrete_passivity}
\end{align}
\begin{tikzpicture}[remember picture, overlay]
    \draw[decorate, decoration={calligraphic brace, amplitude=6pt}, very thick]
        ([xshift=-5pt, yshift=1.5ex]pic cs:ekin_start) --
        ([xshift=-5pt, yshift=-0.5ex]pic cs:ekin_end)
        node[midway, xshift=22pt] {$\Delta V_\mathrm{kin}$};
    \draw[decorate, decoration={calligraphic brace, amplitude=6pt}, very thick]
        ([xshift=-5pt, yshift=1.5ex]pic cs:epot_start) --
        ([xshift=-5pt, yshift=-0.5ex]pic cs:epot_end)
        node[midway, xshift=22pt] {$\Delta V_\mathrm{pot}$};
\end{tikzpicture}
Opposed to energy tank formulations using an energy accumulator state variable, we are using physical spring energy inherent to impedance control.
We thus obtain the next $d_{i,k+1}$
\begin{equation}
	d_{i,k+1} = \sqrt{\frac{P_{\mathrm{d},i,k} \Delta t + \frac{1}{2}d_{i,k}^2\Delta\bm{x}_{i,k}^\top \bm{K}_{\mathrm{p},i,k}^* \Delta\bm{x}_{i,k}}{\frac{1}{2} \Delta\bm{x}_{i,k}^\top \bm{K}_{\mathrm{p},i,k+1}^* \Delta\bm{x}_{i,k} + \epsilon}}. \label{eq:d_evolution}
\end{equation}
%with $\epsilon$ avoiding division by $0$.
%All factors under the square root are positive, leading to a valid result.
%In \Cref{fig:limit_spring_deflection_discrete}, we achieve passivity even for step inputs with time-discrete control.
The approach is summarized in \Cref{algo:limit_spring_deflection_discrete}.
Note that \cite{lee2010passive} arrive at a condition similar to ours in \eqref{eq:spring_deflection:discrete_passivity} treating the different problem of modulating a target for constant stiffness, thus not passivating stiffness changes.

\begin{algorithm}[t]
\caption{\label{algo:limit_spring_deflection_discrete} Limiting the spring deflection at $k+1$.}
\begin{algorithmic}
\State $P_{\mathrm{d}, i, k} \gets \Delta \dot{\bm{x}}_{i,k}^\top \bm{K}_{\mathrm{d},i,k} \Delta \dot{\bm{x}}_{i,k-1}$ \Comment{Damping \eqref{eq:damping_power}}
\State $d_{i,k+1} \gets \mathrm{min}(d_{i,k+1}, 1)$ \Comment{Scaling $d$, limit to $[0, 1]$ \eqref{eq:d_evolution}}
\State $w_{i,k+1} \gets w_{\mathrm{pass}} (d_{i,k+1}, \bm{K}^*_{\mathrm{p},i,k+1}, \bm{K}_{\mathrm{d},i,k+1})$ \Comment{\eqref{eq:impedance_control_pass_l}}
\end{algorithmic}
\end{algorithm}

\subsection{Limiting the Stiffness Change}
\label{sec:stabilization:stiffness_change}
\begin{figure}
	\centering
    \includegraphics[width=0.95\columnwidth,trim={30, 20, 20, 15},clip]{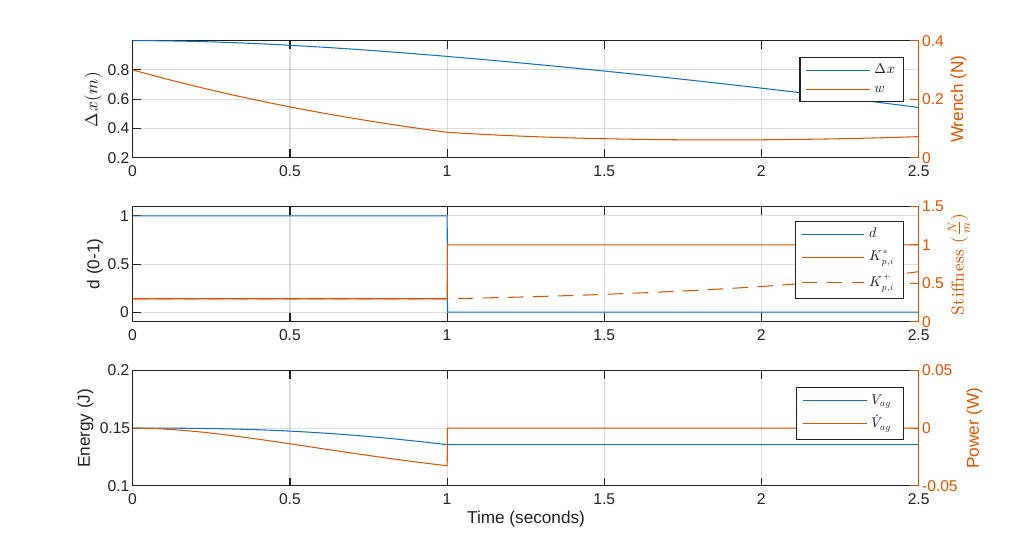}
	\caption{\label{fig:limit_kdot} Limiting the stiffness change rate $\dot{\bm{K}}^+_{\mathrm{p},i}$ to $d_i \dot{\bm{K}}^*_{\mathrm{p},i}$ to ensure passivity.\vspace{-1em}}
\end{figure}
As proposed by \cite{balachandran2022stable,balachandran2023passive}, we also investigate limiting the rate of stiffness change; in our case however for full stiffness matrix changes $\dot{\bm{K}}^*_{\mathrm{p},i}$.
This leads to the storage function
\begin{equation}
	V = \frac{1}{2} \dot{\bm{x}}_\mathrm{ee}^\top \bm{M} \dot{\bm{x}}_\mathrm{ee} + \sum_i \frac{1}{2} \Delta \bm{x}_i^\top \bm{K}^+_{\mathrm{p},i} \Delta \bm{x}_i\label{eq:storage_fcn_k}
\end{equation}
where $\bm{K}^+_{\mathrm{p},i}$ is the stiffness matrix resulting from the modifications of $\bm{K}^*_{\mathrm{p},i}$ by our approach.
%where $\bm{K}^+_{\mathrm{p},i}$ is the stiffness matrix $\bm{K}^*_{\mathrm{p},i}$ modified by our approach.
%The derivative of this storage function evaluates to
Its derivative evaluates to
\begin{align}
	\dot{V} &= \dot{\bm{x}}_\mathrm{ee}^\top \bm{M} \ddot{\bm{x}}_\mathrm{ee} + \sum_i \Delta\bm{x}_i^\top \bm{K}^+_{\mathrm{p},i} \Delta\dot{\bm{x}}_i\nonumber\\
	&+ \sum_i \frac{1}{2} \Delta\bm{x}_i^\top \dot{\bm{K}}^+_{\mathrm{p},i} \Delta\bm{x}_i.\label{eq:power_fcn_k}
\end{align}
As before, we can now in \eqref{eq:power_fcn_k} substitute $\dot{\bm{x}}_\mathrm{ee}^\top \bm{w}_i$ using \eqref{eq:robot_motion} and $\dot{\bm{x}}_i^\top \bm{w}_i = \Delta\dot{\bm{x}}_i^\top \bm{K}^+_{\mathrm{p},i} \Delta \bm{x}_i + \Delta \dot{\bm{x}}_i^\top \bm{K}_{\mathrm{d},i} \Delta \dot{\bm{x}}_i + \dot{\bm{x}}_\mathrm{ee}^\top \bm{w}_i$ to get
\begin{align}
	\dot{V} &= \sum_i \dot{\bm{x}}_i^\top \bm{w}_i + \dot{\bm{x}}_\mathrm{ee}^\top \bm{w}_e - \dot{\bm{x}}_\mathrm{ee}^\top \bm{D} \dot{\bm{x}}_\mathrm{ee} - \sum_i \Delta \dot{\bm{x}}_i^\top \bm{K}_{\mathrm{d},i} \Delta \dot{\bm{x}}_i\nonumber\\
	&+ \sum_i \frac{1}{2} \Delta\bm{x}_i^\top \dot{\bm{K}}^+_{\mathrm{p},i} \Delta\bm{x}_i\label{eq:power_fcn_subst_k}
\end{align}
where the last term possibly violates passivity \eqref{eq:passivity_condition}.
When $\frac{1}{2} \Delta\bm{x}_i^\top \dot{\bm{K}}^+_{\mathrm{p},i} \Delta\bm{x}_i \le 0$, no energy is being generated through the evolution of the stiffness matrix $\bm{K}_{\mathrm{p},i}^*$.
We can then directly set $\bm{K}_{\mathrm{p},i}^+ = \bm{K}_{\mathrm{p},i}^*$ in \eqref{eq:impedance_control_pass_k} and omit limiting the stiffness change through \eqref{eq:stiffness_change:passivity_condition} -- \eqref{eq:stiffness_update}.
For the active case $\frac{1}{2} \Delta\bm{x}_i^\top \dot{\bm{K}}^+_{\mathrm{p},i} \Delta\bm{x}_i > 0$, we set $\dot{\bm{K}}^+_{\mathrm{p},i} = d_i \dot{\bm{K}}^*_{\mathrm{p},i}$, which leads to the passivity condition
\begin{gather}
	\underbrace{P_{\mathrm{d},i}}_{\text{damping power}} \ge \underbrace{\frac{1}{2} \Delta\bm{x}_i^\top d_i \dot{\bm{K}}^*_{\mathrm{p},i} \Delta\bm{x}_i}_{\text{power from $\dot{\bm{K}}^*_{\mathrm{p},i}$ limited by $d_i$}}.\label{eq:stiffness_change:passivity_condition}
\end{gather}
Rearranging terms allows us to calculate $d_i$ through
\begin{equation}
	d_i \le \frac{2 P_{\mathrm{d},i}}{\Delta\bm{x}_i^\top \dot{\bm{K}}^*_{\mathrm{p},i} \Delta\bm{x}_i}.\label{eq:stiffness_change:d}
\end{equation}
Because both $P_{\mathrm{d},i} \ge 0$ as well as $\Delta\bm{x}_i^\top \dot{\bm{K}}^*_{\mathrm{p},i} \Delta\bm{x}_i > 0$ in the active case, $d_i \ge 0$ holds.
We subsequently limit $d_i \le 1$ to ensure that the updated stiffness never exceeds nominal $\bm{K}^*_{\mathrm{p},i}$.

In a discrete time controller, we calculate the next stiffness
\begin{equation}
	\bm{K}^+_{\mathrm{p},i,k+1} = \bm{K}^+_{\mathrm{p},i,k} + d_{i,k+1} \dot{\bm{K}}^*_{\mathrm{p},i,k+1} \Delta t
\end{equation}
using the unfiltered derivative
\begin{equation}
\dot{\bm{K}}^*_{\mathrm{p},i,k+1} = \frac{\bm{K}^*_{\mathrm{p},i,k+1} - \bm{K}^+_{\mathrm{p},i,k}}{\Delta t}.\label{eq:stiffness_change:derivative}
\end{equation}
This leads to the update formula
\begin{equation}
	\bm{K}^+_{\mathrm{p},i,k+1} = d_{i,k+1} \bm{K}^*_{\mathrm{p},i,k+1} + (1-d_{i,k+1}) \bm{K}^+_{\mathrm{p},i,k} \label{eq:stiffness_update}
\end{equation}
with $\bm{K}^+_{\mathrm{p},i,k+1}$ positive semidefinite as $d_{i,k+1} \in [0, 1]$ and both $\bm{K}^*_{\mathrm{p},i,k+1}$ and $\bm{K}^+_{\mathrm{p},i,1}$ are positive semidefinite matrices \cite{horn2017matrix}.

Finally, the impedance control law \eqref{eq:impedance_control} is updated with $\bm{K}^+_{\mathrm{p},i}$
\begin{equation}
	\bm{w}_{\mathrm{pass},i} = \bm{K}^+_{\mathrm{p},i} \mathrm{Log}_{\bm{x}_{\mathrm{ee}}}\left(\bm{x}_i\right) + \bm{K}_{\mathrm{d},i} \Delta \dot{\bm{x}}_i,\label{eq:impedance_control_pass_k}
\end{equation}
resulting in a passivated wrench $\bm{w}_{\mathrm{pass},i}$.
The approach is summarized in \Cref{algo:limit_stiffness}.
\Cref{fig:limit_kdot} shows the result of using this approach with the same stiffness step input as before.
%Like for the time-discrete deflection limitation shown in \Cref{fig:limit_spring_deflection_discrete}, no passivity violation occurs.

\begin{algorithm}[t]
\caption{\label{algo:limit_stiffness} Limiting the stiffness change at $k+1$.}
\begin{algorithmic}
\State $P_{\mathrm{d}, i, k+1} \gets \Delta \dot{\bm{x}}_{i,k}^\top \bm{K}_{\mathrm{d},i,k+1} \Delta \dot{\bm{x}}_{i,k}$ \Comment{Damping \eqref{eq:damping_power}}
\State $P_{\mathrm{a}, i, k+1} \gets \frac{1}{2} \Delta\bm{x}_{i,k}^\top \dot{\bm{K}}^*_{\mathrm{p},i,k+1} \Delta\bm{x}_{i,k}$ \Comment{\eqref{eq:stiffness_change:passivity_condition} w. der. \eqref{eq:stiffness_change:derivative}}
\If{$P_{\mathrm{a},i,k+1} > 0$} \Comment{Stiffness change is active}
    \State $d_{i, k+1} \gets min(d_{i,k+1}, 1)$ \Comment{Limit $d$ to $[0, 1]$ \eqref{eq:stiffness_change:d}}
\Else
    \State $d_{i, k+1} \gets 1$ \Comment{Stiffness constant or decreasing}
\EndIf
\State $\bm{K}^+_{\mathrm{p},i,k+1} \gets \bm{K}^+_{\mathrm{p},i,k+1} (\bm{K}^*_{\mathrm{p},i,k+1}, \bm{K}^+_{\mathrm{p},i,k})$ \Comment{Stiffness \eqref{eq:stiffness_update}}
\State $w_{i,k+1} \gets w_{\mathrm{pass}} (\bm{K}^+_{\mathrm{p},i,k+1}, \bm{K}_{\mathrm{d},i,k+1})$ \Comment{Wrench \eqref{eq:impedance_control_pass_k}}
\end{algorithmic}
\end{algorithm}

\subsection{Extensions of the Passivation Approaches}
\label{sec:stabilization:extensions}
\subsubsection{Passivating Asymmetric Stiffness Elements}
Both when limiting the spring deflection and when limiting the stiffness change, for $d_i = 1$, the derivatives of the storage function \eqref{eq:power_fcn} and \eqref{eq:power_fcn_k} have possibly passivity-violating powers from the evolution of the symmetric part of the stiffness matrix $P_{\mathrm{a}, i}$ as well as from the asymmetric part $P_{\mathrm{curl},i}$ (cf. \cite{tsuji2023stability})
\begin{gather}
	P_{\mathrm{a}, i} = \frac{1}{2} \Delta\bm{x}_i^\top \dot{\bm{K}}^*_{\mathrm{p},i} \Delta\bm{x}_i, \quad P_{\mathrm{curl},i} = -\Delta \dot{\bm{x}}_i^\top \bm{K}_{\mathrm{p},\mathrm{curl},i}^{*} \Delta \bm{x}_i.\nonumber
\end{gather}
When the damping $P_{\mathrm{d}, i}$ \eqref{eq:damping_power} is larger than the active power $P_{\mathrm{a}, i}$ of the change of symmetric part $\bm{K}_{\mathrm{p},i}^*$ passivated in \Cref{sec:stabilization:spring_deflection,sec:stabilization:stiffness_change}, we can calculate
\begin{gather}
	d_{\mathrm{curl},i} = \mathrm{min} \left( \frac{P_{\mathrm{d},i} - P_{\mathrm{a},i}}{P_{\mathrm{curl},i}}, 1 \right), \quad \bm{w}_{\mathrm{curl},i} = d_{\mathrm{curl},i} \bm{K}_{\mathrm{p},\mathrm{curl},i}^{*} \Delta \bm{x}_i\nonumber
\end{gather}
which is then added to the passivated wrenches \eqref{eq:impedance_control_pass_l} or \eqref{eq:impedance_control_pass_k}.
We evaluate such stiffness matrices in \Cref{sec:eval:asym_stiff}.

\subsubsection{Initial Energy}
\label{sec:stabilization:extensions_initial}
Starting with $\dot{\bm{x}}_i = \bm{0}$ from $\bm{K}_{\mathrm{p},i}^* = \bm{0}$ and $\dot{\bm{x}}_\mathrm{ee} = \bm{0}$ in autonomous systems, the robot will not move as no damping power is available to achieve a stiffness increase.
Similar to energy tanks, we can equip the system with some initial energy $V_\mathrm{init}$ that can be used at rate $\dot{V}_\mathrm{init} \le P_\mathrm{init}$, thus leading to a modified $P_{\mathrm{d},i}' = P_{\mathrm{d},i} + \dot{V}_\mathrm{init}$ until $V_\mathrm{init} = 0$.
We evaluate an autonomous system utilizing this initial energy in \Cref{sec:eval:rotations_impacts}.
Feeding $P_{\mathrm{d},i} - P_{\mathrm{a},i} - P_{\mathrm{curl},i}$ back into $V_\mathrm{init}$ would result in a full energy tank, however for different passivity condition than \eqref{eq:passivity_condition} and potential energy accumulation.

%%%%%%%%%%%%%%%%%%%%%%%%%%%%%%%%%%%%%%%%%%%%%%%%%%%%%%%%%%%%%%%%%%%%%%%%%%%%%%%%
\section{Evaluation}
\label{sec:experiments}
\begin{figure}
	\centering
    \includegraphics[width=0.8\columnwidth,trim={30, 20, 20, 15},clip]{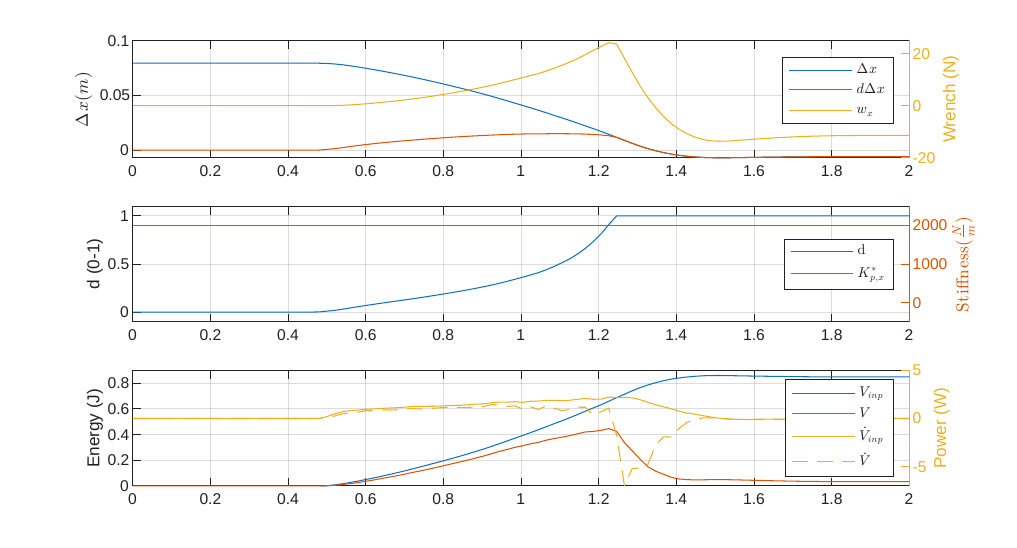}
	\caption{\label{fig:limit_spring_deflection_real} Initializing the approach of limiting the spring deflection. An external wrench initializing the movement is applied at $t=\SI{0.5}{\second}$.\vspace{-1em}}
\end{figure}
\begin{figure}
	\centering
    \includegraphics[width=0.8\columnwidth,trim={30, 20, 20, 15},clip]{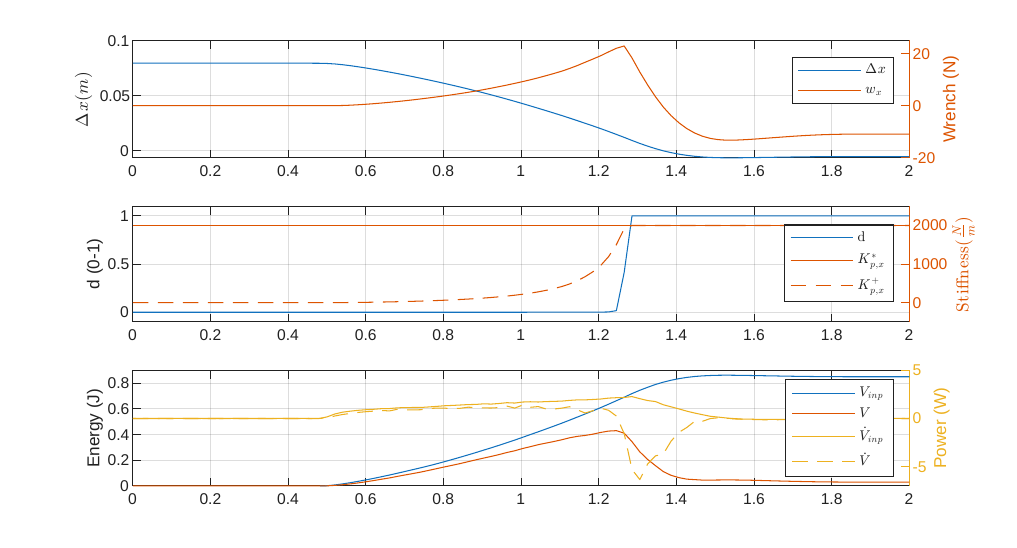}
	\caption{\label{fig:limit_kdot_real} Initializing the approach of limiting the stiffness change. An external wrench initializing the movement is applied at $t=\SI{0.5}{\second}$.\vspace{-1em}}
\end{figure}

We evaluate the proposed passivation approaches on 7-\gls{dof} lightweight robot systems both quantitatively using precoded input wrenches simulating user interaction as well as qualitatively using real user input and shared control methods~\cite{muehlbauer2024probabilistic,muehlbauer2025unified}.
The supplementary video shows a visualization of the experiments.
We use double diagonalization~\cite{albuschaeffer2003cartesianimpedance} with desired stiffness as input to compute the damping matrix.
In \eqref{eq:d_evolution}, we set $\epsilon$ to the machine epsilon.
The robot used in \Cref{sec:eval:advanced} is controlled at \SI{1}{\kilo\hertz} with a lowpass filter on the measured velocities with cutoff frequency of \SI{6}{\hertz} while the robot used in other sections is controlled at \SI{8}{\kilo\hertz} with velocities filtered at \SI{100}{\hertz}.
Velocities of attractor poses are filtered at \SI{10}{\hertz}.
No filtering is applied to the calculated wrenches / torques.

\subsection{Method Initialization}
\label{sec:eval:init}
Steps in the desired stiffness values happen when enabling a method or the robot.
Initializing the springs with $\bm{K}_{\mathrm{p},i}^* = \bm{0}$ thus ensures an understandable behavior of the robotic system, as the robot will only start to move through user interaction.
%Full stiffness and therefore nominal behavior is still reached very quickly as can be seen in \Cref{fig:limit_spring_deflection_real,fig:limit_kdot_real}.
\Cref{fig:limit_spring_deflection_real,fig:limit_kdot_real} show the initialization of both approaches, where the robot initially does not move, even with a non-zero stiffness.
The movement only starts when an external wrench $\bm{w}_\mathrm{ext} = (\SI{10}{\newton}, 0, 0, 0, 0, 0)^\top$ - which can be compared to a user interaction - is applied at $t=\SI{0.5}{\second}$.
Both methods ensure that the energy $V$ stored in the system according to \eqref{eq:storage_fcn} respectively \eqref{eq:storage_fcn_k} is lower than the energy supplied at the input $V_\mathrm{inp}$, which can be seen through $\dot{V} \le \dot{V}_\mathrm{inp}$.
They achieve a very similar behavior in this initialization setting.

\subsection{Rotations and Impacts}
\label{sec:eval:rotations_impacts}
\begin{figure}
	\centering
    \includegraphics[width=0.8\columnwidth,trim={30, 20, 20, 15},clip]{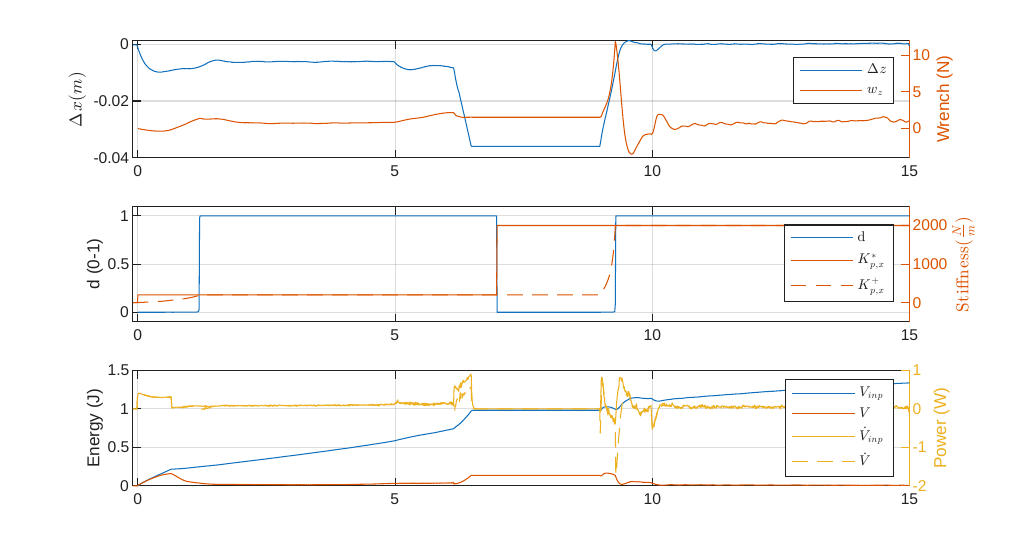}
	\caption{\label{fig:limit_kdot_impact} Testing the approach of limiting the stiffness change with rotational movements and interaction with a hard environment.\vspace{-1em}}
\end{figure}
In a setting with initial orientation and position offset, an \textit{initial energy} (\Cref{sec:stabilization:extensions}) of $\SI{0.2}{\joule}$ is used at a rate of $\SI{0.3}{\watt}$ to start robot motion when activating an attractor moving both in $z$ direction and in its orientation (\Cref{fig:limit_kdot_impact}).
At $t=\SI{6}{\second}$, the robot comes in contact with a hard environment.
Until $t=\SI{6.5}{\second}$, the attractor moves further away in direction of the blocked motion, elongating the spring and introducing energy in the system using the power port $(\dot{\bm{x}}_i, \bm{w}_i)$.
At $t=\SI{7}{\second}$, a jump in the desired stiffness is commanded which is not realized as with $\Delta\dot{\bm{x}}_i = \bm{0}$, no damping energy is available.
Only after $t=\SI{9}{\second}$, the desired stiffness is realized gradually in a passivity-preserving manner, as can be seen through $\dot{V} < \dot{V}_\mathrm{inp}$.

Over $10$ repetitions, we see passivity violations of more than \SI{0.02}{\watt} ($\dot{V} > \dot{V}_\mathrm{inp} + \SI{0.02}{\watt}$) for \SI{0.83 \pm 0.32}{\percent} (spring deflection limiting) respectively \SI{0.73 \pm 0.34}{\percent} (stiffness change limiting) of time steps.
The total energy introduced in the system through every passivity violation ($\dot{V} > \dot{V}_\mathrm{inp}$) sums to \SI{0.016 \pm 0.004}{\joule} (spring deflection limiting) respectively \SI{0.015 \pm 0.006}{\joule} (stiffness change limiting).
Note that most of these passivity violations can be attributed to sensor noise, e.g., in the filtered velocity signal.
Compared to an execution without passivation of the spring with passivity violations at \SI{1.78 \pm 0.33}{\percent} of the time steps and an energy of \SI{1.75 \pm 0.36}{\joule} introduced through non-passive stiffness changes, we achieve a near-perfect passivation according to our power ports and without storage function.

As can be seen in \Cref{fig:limit_kdot_impact}, the input energy $V_\mathrm{inp}$ accumulates substantially over time.
Thus, even when mitigating shortcomings of the energy tank approach evaluated in the previous section, this energy when stored in a tank could lead to some quite violent and unexpected robot motions at a later point.

\subsection{Comparison with Energy Tanks and TDPA}
\label{sec:eval:energy_tanks}
\begin{figure}
	\centering
    \includegraphics[width=0.8\columnwidth,trim={30, 10, 20, 10},clip]{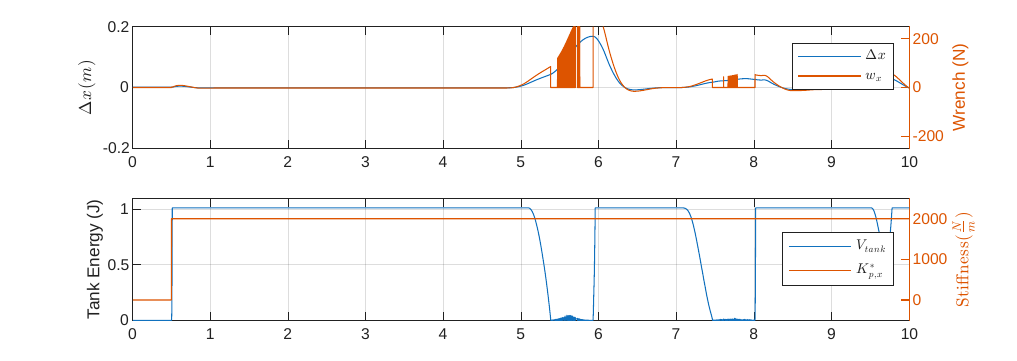}
	\caption{\label{fig:baseline_tank} At $t=\SI{0.5}{\second}$, an attractor is commanded to the baseline~\cite{ferraguti2013tank}. Due to the tank containing more energy than its lower limit $\epsilon$, the stiffness change is realized instantly, leading to a fast robot motion. At $t=\SI{5}{\second}$, a human operator starts deflecting the spring.
As soon as the tank is depleted, chattering leads to the spring being activated and deactivated at high frequency.\vspace{-1.5em}}
\end{figure}
We implement the energy-tank based approach \cite{ferraguti2013tank} as baseline with $\bar{T} = \SI{1}{\joule}$ and $\epsilon = \SI{0.01}{\joule}$.
Figure~\ref{fig:baseline_tank} shows the behavior when initializing and during human interaction - we observe fast motion on initialization as well as chattering when the spring is deflected.
In contrast, our approach of limiting the spring deflection only fades in the spring  with energy provided by the human operator.
Subsequently, nominal spring behavior is retained as seen in the supplementary video.

For both energy tanks and TDPA~\cite{hannaford2002time} we repeat the experiments from the two previous sections.
We observe passivity violations of \SI{0.036}{\joule} (energy tank) and \SI{0.033}{\joule} (TDPA) compared to \SI{0.002}{\joule} for the initialization (\Cref{sec:eval:init}).
For the impact experiment, passivity violations of \SI{1.17}{\joule} (energy tank) and \SI{1.01}{\joule} (TDPA), which are considerably larger than the values observed for our method (\Cref{sec:eval:rotations_impacts}).
Robot behavior, such as fast motion and chattering can be seen in the video.

\subsection{Spring Deflection vs. Stiffness Change Limitation}
\label{sec:eval:comp}
\begin{figure}
	\centering
    \includegraphics[width=0.8\columnwidth,trim={30, 20, 20, 15},clip]{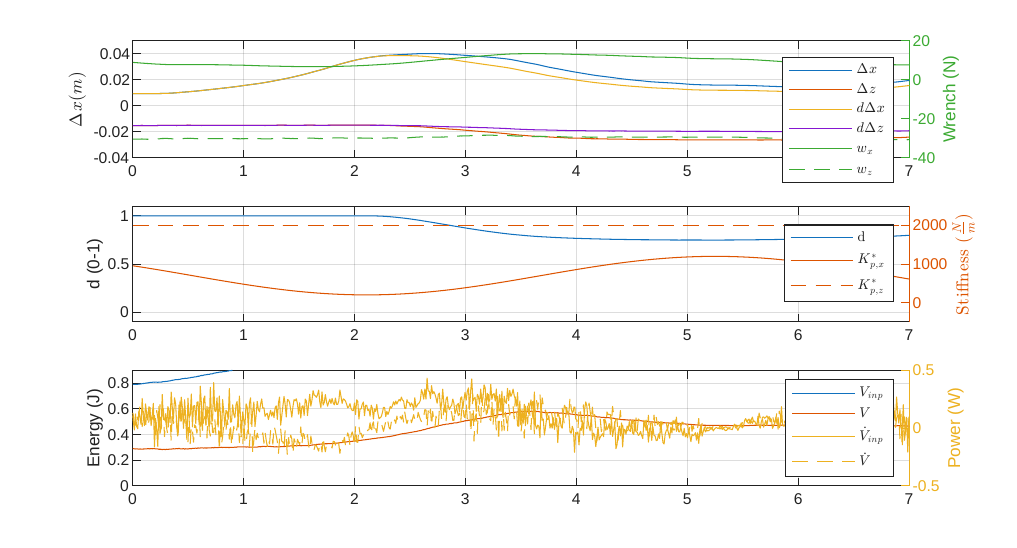}
	\caption{\label{fig:limit_spring_deflection_real_trajectory} Moving the robot along the $y$-axis while modulating the stiffness of the $x$-axis with external forces applied. Through the lever-based limiting of the spring deflection, the error $\Delta z$ is also increasing with decreasing $d$.}
\end{figure}
\begin{figure}
	\centering
    \includegraphics[width=0.8\columnwidth,trim={30, 20, 20, 15},clip]{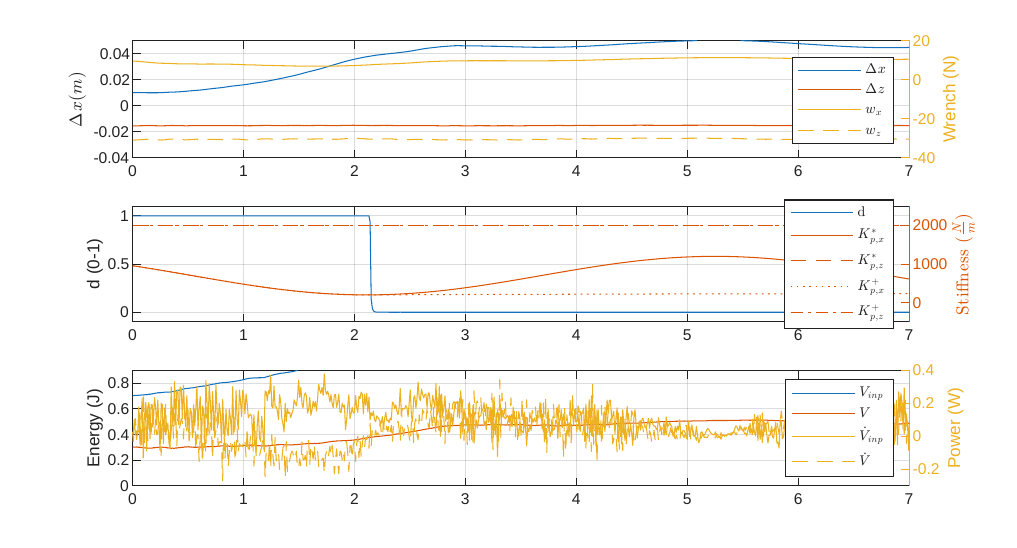}
	\caption{\label{fig:limit_kdot_real_trajectory}  Moving the robot along the $y$-axis while modulating the stiffness of the $x$-axis with external forces applied. Limiting the change rate of $\bm{K}$, the error $\Delta z$ does not change as the stiffness in this direction is constant.\vspace{-1em}}
\end{figure}

To explore the different behaviors of both passivation approaches, we set up a demonstration scenario where the attractor point is moving for \SI{20}{\centi\metre} along the $y$-axis with changing stiffness along the $x$-axis.
In particular, we apply the stiffness $\bm{K} = \mathrm{diag}\left( 700 + 500 \cdot \mathrm{sin}\left( t \right), 2000, 2000, 50, 50, 50.0 \right) \mathrm{\frac{N}{m}}$ and an external wrench $\bm{w}_\mathrm{ext} = (10, 0, -30, 0, 0, 0)^\top \mathrm{N}$.
As can be seen in \Cref{fig:limit_spring_deflection_real_trajectory}, the stiffness varying in $x$ also affects the error $\Delta z$ of the $z$-axis along which an external force is being applied as the controller is scaled by the scalar value $d$.
In contrast, when limiting the stiffness change (\Cref{fig:limit_kdot_real_trajectory}), the error $\Delta z$ stays constant as this method only affects the stiffness along the $x$-axis while the stiffness along the $z$-axis remains constant.

\subsection{Evaluation with Coupled and Geometric Stiffness Matrices}
\label{sec:eval:advanced}
\begin{figure}
	\centering
    \includegraphics[width=0.8\columnwidth,trim={30, 5, 20, 10},clip]{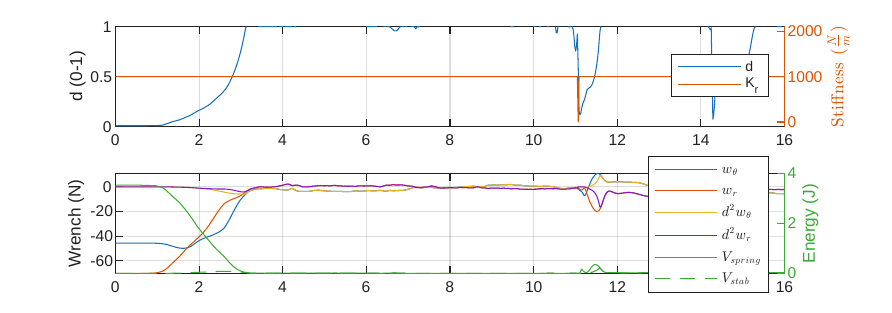}
	\caption{\label{fig:coupled_stiffness} In this scenario, a human operator starts the robot in a configuration with a deflected spring ($V_\mathrm{spring} > 0$). Thanks to limiting the spring deflection, the spring is only activated as the operator moves.\vspace{-1em}}
\end{figure}
\begin{figure}
	\centering
    \includegraphics[width=0.8\columnwidth,trim={30, 5, 20, 10},clip]{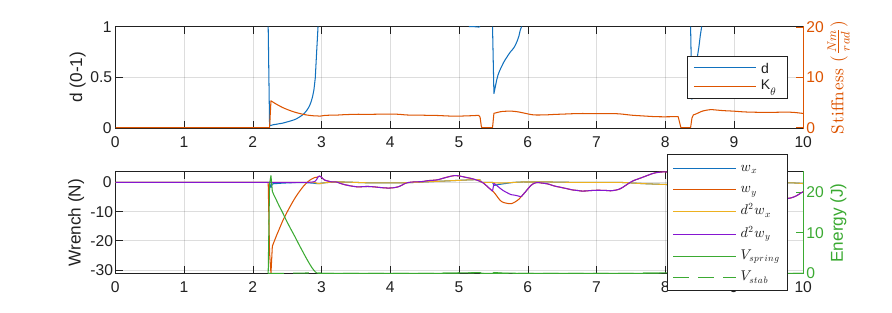}
	\caption{\label{fig:ring_tubes} Passivating an attractor on the cylindrical manifold $\mathcal{M}_2$ of~\cite{muehlbauer2025unified}. An attractor gets activated when the operator moves closer to it while limiting the spring deflection ensures a gradual enabling of the spring.\vspace{-1em}}
\end{figure}
We demonstrate how two selected shared control scenarios can be passivated using our method.
\Cref{fig:coupled_stiffness} shows how the energy stored in the springs evolves for a complex scenario with a single attractor and stiffness matrices coupling positional and orientational degrees of freedom.
\glsreset{vf}
In \Cref{fig:ring_tubes}, a step response is visualized when a single attractor on a cylindrical manifold is activated abruptly.
See the supplementary video for a visualization of the interaction with a human user.

\subsection{Passivating the Arbitration of Multiple Attractors}
\label{sec:eval:switching}
\begin{figure}
	\centering
    \includegraphics[width=0.8\columnwidth,trim={30, 10, 20, 10},clip]{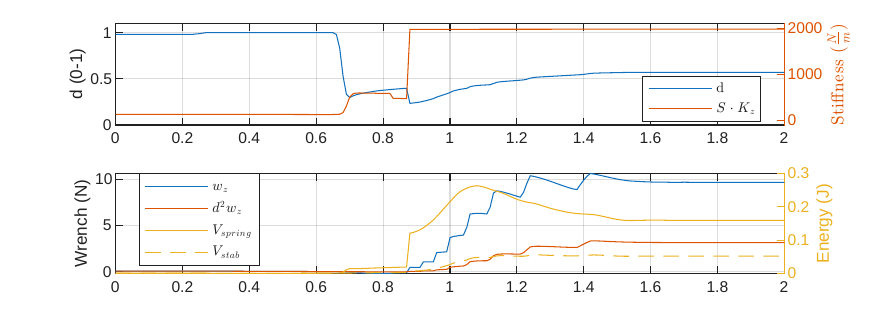}
	\caption{\label{fig:hug_limit_spring_deflection_discrete} Passivation of the arbitration \eqref{eq:impedance_to_torque} using the stiffness adaptation of \Cref{sec:arbitration_as_modulation}, modifying the nominal $K_z = \SI{2000}{\frac{N}{m}}$ of the second trajectory. Reducing the stiffness of the first trajectory is passive and thus not shown here.}
\end{figure}
The transition between two trajectories, where the attractors are selected as the closest point of the trajectory to the end effector, is passivated through reformulating the arbitration as stiffness modulation (\Cref{sec:arbitration_as_modulation}).
While the first trajectory is equipped with variable stiffness, the second is using a constant stiffness which is only modulated by our arbitration.
\Cref{fig:hug_limit_spring_deflection_discrete} shows the behavior of the system, applying a wrench of \SI{-10}{\newton} along the $z$-axis to simulate user input.
Limiting the discrete spring deflection, a smooth transition is ensured.

\subsection{Asymmetric Stiffnesses}
\label{sec:eval:asym_stiff}
Two attractors arbitrated by their covariance as in~\cite{muehlbauer2025unified} result in asymmetric entries of the stiffness matrix.
After reaching the final attractor, $P_{\mathrm{curl},i}$ decreases and $d_{\mathrm{curl},i}$ approaches $1$.
We allow for an active power of $P_{\mathrm{curl},i} < \SI{0.1}{\watt}$ to realize the asymmetric component also in cases of velocity measurement noise.
See the supplementary video for details.

In this case, a Gaussian product would yield an attractor corresponding to the solution with asymmetric stiffness components.
When utilizing different manifolds $\mathcal{M}$ for individual attractors or different methods of generating wrenches, such straightforward solution is not possible, highlighting the importance of allowing for an asymmetric stiffness component.

%%%%%%%%%%%%%%%%%%%%%%%%%%%%%%%%%%%%%%%%%%%%%%%%%%%%%%%%%%%%%%%%%%%%%%%%%%%%%%%%
\section{Discussion}
\label{sec:discussion}
Both methods achieve a passivation of shared control approaches using variable impedance.
In particular, attractors can be initialized with high stiffness even when the robot is currently far away from their pose (\Cref{sec:eval:init,sec:eval:advanced}).
Through reformulation of the arbitration, a stable switching between fixtures can be achieved as shown in \Cref{sec:eval:switching}.

Differences between both approaches become apparent when the variable stiffness only affects single \glspl{dof} (\Cref{sec:eval:comp}).
Limiting the spring deflection scales the whole deflection with a scalar, thus preserving the \textit{characteristics} of the stiffness.
This, however, might also lead to a downscaling of already stiffened up \glspl{dof}.
In contrast, the approach of limiting the stiffness change does not decrease the stiffness of \glspl{dof} where the stiffness matrix does not change.
This might, however, change the overall impedance characteristic as the ratio between individual \glspl{dof} is not preserved.

\textbf{Choosing the passivation method:}
\begin{itemize}
	\item \textit{uniformly varying} stiffness: either method is applicable,
	\item \textit{keep characteristics} of the stiffness matrix: limit the spring deflection (\Cref{algo:limit_spring_deflection_discrete}),
	\item \textit{keep stiffness} in non-changing \glspl{dof}: limit the stiffness change (\Cref{algo:limit_stiffness}).
\end{itemize}

%%%%%%%%%%%%%%%%%%%%%%%%%%%%%%%%%%%%%%%%%%%%%%%%%%%%%%%%%%%%%%%%%%%%%%%%%%%%%%%%
\section{Conclusion}
We have presented a novel approach to the passivation of shared control methods using variable impedance supporting an arbitrary number of concurrently active controllers.
We have reformulated the arbitration of individual controllers as stiffness modulation.
Following a passivity analysis both of this arbitration as well as of variable impedance, two approaches for passivization are proposed.
Based on an analysis of the storage function both when limiting the spring deflection as well as when limiting the stiffness matrix change rate, necessary conditions for achieving passivity are given.
Furthermore, the different characteristics of both methods are evaluated.
The resulting methods can be used to passivate standard impedance controllers and allow for a stabilization of shared control methods without posing constraints on the evolution of stiffness matrices, initialization of attractors or arbitration.
With this gained flexibility, we hope to stimulate shared control research.
Future work should also extend our approach to the general variable impedance setting, e.g., by using power inputs different to the human operator or moving attractor.

%%%%%%%%%%%%%%%%%%%%%%%%%%%%%%%%%%%%%%%%%%%%%%%%%%%%%%%%%%%%%%%%%%%%%%%%%%%%%%%%
%\section*{Acknowledgments}
%\label{sec:acknowledgements}

%Add funding agencies here.

%endispell

\bibliographystyle{./style/IEEEtran}
\bibliography{literatur}

\begin{thebibliography}{10}
\providecommand{\url}[1]{#1}
\csname url@samestyle\endcsname
\providecommand{\newblock}{\relax}
\providecommand{\bibinfo}[2]{#2}
\providecommand{\BIBentrySTDinterwordspacing}{\spaceskip=0pt\relax}
\providecommand{\BIBentryALTinterwordstretchfactor}{4}
\providecommand{\BIBentryALTinterwordspacing}{\spaceskip=\fontdimen2\font plus
\BIBentryALTinterwordstretchfactor\fontdimen3\font minus
  \fontdimen4\font\relax}
\providecommand{\BIBforeignlanguage}[2]{{%
\expandafter\ifx\csname l@#1\endcsname\relax
\typeout{** WARNING: IEEEtran.bst: No hyphenation pattern has been}%
\typeout{** loaded for the language `#1'. Using the pattern for}%
\typeout{** the default language instead.}%
\else
\language=\csname l@#1\endcsname
\fi
#2}}
\providecommand{\BIBdecl}{\relax}
\BIBdecl

\bibitem{hogan1984impedance}
N.~Hogan, ``Impedance control: An approach to manipulation,'' in \emph{1984
  American Control Conference}.\hskip 1em plus 0.5em minus 0.4em\relax IEEE, 07
  1984.

\bibitem{abifarraj2017learning}
F.~Abi-Farraj, T.~Osa, N.~P.~J. Peters, G.~Neumann, and P.~R. Giordano, ``A
  learning-based shared control architecture for interactive task execution,''
  in \emph{2017 IEEE International Conference on Robotics and Automation
  (ICRA)}.\hskip 1em plus 0.5em minus 0.4em\relax IEEE, 05 2017.

\bibitem{muehlbauer2025unified}
\BIBentryALTinterwordspacing
M.~Mühlbauer, B.~Weber, S.~Calinon, F.~Stulp, A.~Albu-Schäffer, and
  J.~Silvério, ``A unified framework for probabilistic dynamic-, trajectory-
  and vision-based virtual fixtures,'' 2025. [Online]. Available:
  \url{https://arxiv.org/abs/2506.10239}
\BIBentrySTDinterwordspacing

\bibitem{abbink2011haptic}
D.~A. Abbink, M.~Mulder, and E.~R. Boer, ``Haptic shared control: smoothly
  shifting control authority?'' \emph{Cognition, Technology \& Work}, vol.~14,
  no.~1, pp. 19--28, 2011.

\bibitem{abudakka2020variable}
F.~J. Abu-Dakka and M.~Saveriano, ``Variable impedance control and learning—a
  review,'' \emph{Frontiers in Robotics and AI}, vol.~7, 2020.

\bibitem{franzese2021ilosa}
G.~Franzese, A.~Meszaros, L.~Peternel, and J.~Kober, ``Ilosa: Interactive
  learning of stiffness and attractors,'' in \emph{2021 IEEE/RSJ International
  Conference on Intelligent Robots and Systems (IROS)}.\hskip 1em plus 0.5em
  minus 0.4em\relax IEEE, 09 2021, pp. 7778--7785.

\bibitem{muehlbauer2024probabilistic}
M.~Mühlbauer, T.~Hulin, B.~Weber, S.~Calinon, F.~Stulp, A.~Albu-Schäffer, and
  J.~Silvério, ``A probabilistic approach to multi-modal adaptive virtual
  fixtures,'' \emph{IEEE Robotics and Automation Letters}, vol.~9, no.~6, pp.
  5298--5305, 2024.

\bibitem{michel2023learning}
Y.~Michel, Z.~Li, and D.~Lee, ``A learning-based shared control approach for
  contact tasks,'' \emph{IEEE Robotics and Automation Letters}, vol.~8, no.~12,
  pp. 8002--8009, 2023.

\bibitem{ferraguti2013tank}
F.~Ferraguti, C.~Secchi, and C.~Fantuzzi, ``A tank-based approach to impedance
  control with variable stiffness,'' in \emph{2013 IEEE International
  Conference on Robotics and Automation}.\hskip 1em plus 0.5em minus
  0.4em\relax IEEE, 2013, pp. 4948--4953.

\bibitem{hannaford2002time}
B.~Hannaford and J.-H. Ryu, ``Time-domain passivity control of haptic
  interfaces,'' \emph{IEEE Transactions on Robotics and Automation}, vol.~18,
  no.~1, pp. 1--10, 2002.

\bibitem{kronander2016stability}
K.~Kronander and A.~Billard, ``Stability considerations for variable impedance
  control,'' \emph{IEEE Transactions on Robotics}, vol.~32, no.~5, pp.
  1298--1305, 2016.

\bibitem{bednarczyk2020passivity}
M.~Bednarczyk, H.~Omran, and B.~Bayle, ``Passivity filter for variable
  impedance control,'' in \emph{2020 IEEE/RSJ International Conference on
  Intelligent Robots and Systems (IROS)}.\hskip 1em plus 0.5em minus
  0.4em\relax IEEE, 2020, pp. 7159--7164.

\bibitem{balachandran2021finite}
R.~Balachandran, H.~Mishra, M.~Panzirsch, and C.~Ott, ``A finite-gain stable
  multi-agent robot control framework with adaptive authority allocation,'' in
  \emph{2021 IEEE International Conference on Robotics and Automation
  (ICRA)}.\hskip 1em plus 0.5em minus 0.4em\relax IEEE, 2021, pp. 1579--1585.

\bibitem{balachandran2023passive}
R.~Balachandran, M.~De~Stefano, H.~Mishra, C.~Ott, and A.~Albu-Schäffer,
  ``Passive arbitration in adaptive shared control of robots with variable
  force and stiffness scaling,'' \emph{Mechatronics}, vol.~90, p. 102930, 2023.

\bibitem{khalil2002nonlinear}
H.~K. Khalil, \emph{Nonlinear Systems}.\hskip 1em plus 0.5em minus 0.4em\relax
  Prentice Hall, 2002, vol.~3.

\bibitem{ti2023geometric}
B.~Ti, A.~Razmjoo, Y.~Gao, J.~Zhao, and S.~Calinon, ``A geometric optimal
  control approach for imitation and generalization of manipulation skills,''
  \emph{Robotics and Autonomous Systems}, vol. 164, p. 104413, 2023.

\bibitem{dyck2022impedance}
M.~Dyck, A.~Sachtler, J.~Klodmann, and A.~Albu-Schaffer, ``Impedance control on
  arbitrary surfaces for ultrasound scanning using discrete differential
  geometry,'' \emph{IEEE Robotics and Automation Letters}, vol.~7, no.~3, pp.
  7738--7746, 2022.

\bibitem{albuschaeffer2003cartesianimpedance}
A.~Albu-Schäffer, C.~Ott, U.~Frese, and G.~Hirzinger, ``Cartesian impedance
  control of redundant robots: recent results with the dlr-light-weight-arms,''
  in \emph{2003 IEEE International Conference on Robotics and Automation},
  vol.~3, 2003, pp. 3704--3709 vol.3.

\bibitem{fasse1997spatial}
E.~D. Fasse, ``On the spatial compliance of robotic manipulators,''
  \emph{Journal of Dynamic Systems, Measurement, and Control}, vol. 119, no.~4,
  pp. 839--844, 1997.

\bibitem{zhang2000spatial}
S.~Zhang and E.~D. Fasse, ``Spatial compliance modeling using a
  quaternion-based potential function method,'' \emph{Multibody System
  Dynamics}, vol.~4, no.~1, pp. 75--101, 2000.

\bibitem{raiola2017comanipulation}
G.~Raiola, S.~S. Restrepo, P.~Chevalier, P.~Rodriguez-Ayerbe, X.~Lamy,
  S.~Tliba, and F.~Stulp, ``Co-manipulation with a library of virtual guiding
  fixtures,'' \emph{Autonomous Robots}, vol.~42, no.~5, pp. 1037--1051, 11
  2017.

\bibitem{selvaggio2016enhancing}
M.~Selvaggio, G.~Notomista, F.~Chen, B.~Gao, F.~Trapani, and D.~Caldwell,
  ``Enhancing bilateral teleoperation using camera-based online virtual
  fixtures generation,'' in \emph{2016 IEEE/RSJ International Conference on
  Intelligent Robots and Systems (IROS)}.\hskip 1em plus 0.5em minus
  0.4em\relax IEEE, 10 2016.

\bibitem{balachandran2022stable}
\BIBentryALTinterwordspacing
R.~Balachandran, ``A stable and transparent framework for adaptive shared
  control of robots,'' Ph.D. dissertation, Technische Universitaet Muenchen,
  2022. [Online]. Available: \url{https://elib.dlr.de/188393/}
\BIBentrySTDinterwordspacing

\bibitem{featherstone2008dynamics}
R.~Featherstone and D.~E. Orin, ``Dynamics,'' in \emph{Springer Handbook of
  Robotics}.\hskip 1em plus 0.5em minus 0.4em\relax Springer Berlin Heidelberg,
  2008, pp. 35--65.

\bibitem{lee2010passive}
D.~Lee and K.~Huang, ``Passive-set-position-modulation framework for
  interactive robotic systems,'' vol.~26, no.~2, pp. 354--369.

\bibitem{horn2017matrix}
R.~A. Horn, \emph{Matrix analysis}, 2nd~ed., C.~R. Johnson, Ed.\hskip 1em plus
  0.5em minus 0.4em\relax Cambridge University Press, 2017.

\bibitem{tsuji2023stability}
T.~Tsuji and Y.~Kato, ``Stability analysis of admittance control using
  asymmetric stiffness matrix,'' 2023.

\end{thebibliography}

\end{document}